%% file: main.tex
\documentclass[11pt]{article}
\usepackage[margin=1in]{geometry}
\usepackage[authoryear,round]{natbib}
\usepackage{times}
\usepackage{microtype}
\usepackage{amsmath,amssymb}
\usepackage{graphicx}
\usepackage{booktabs}
\usepackage{xcolor}
\usepackage{hyperref}
\usepackage{url}
\definecolor{darkblue}{rgb}{0, 0, 0.5}
\hypersetup{colorlinks=true, citecolor=darkblue, linkcolor=darkblue, urlcolor=darkblue}
\graphicspath{{figs/}}

\newcommand{\Es}{\textsc{ES}}
\newcommand{\Pg}{\textsc{PG}}
\newcommand{\Teff}{\ensuremath{T_{\mathrm{eff}}}}
\newcommand{\Deff}{\ensuremath{D_{\mathrm{eff}}}}

\title{Dimension Is Free, Horizon Is Not:\\
When Evolution Strategies Rival Policy Gradients\\
for LLM Fine-Tuning}

\author{%
Qiyao Liang$^{1,2,}$\thanks{\texttt{qiyao@mit.edu}. Work done during an internship at Cognizant AI Lab. A preliminary version of this work appeared at the SciFM workshop at COLM 2026.}\qquad
Kajetan Schweighofer$^{1}$\qquad
Xin Qiu$^{1}$\qquad
Risto Miikkulainen$^{1,3}$\\[6pt]
{\normalsize $^{1}$Cognizant AI Lab\qquad $^{2}$MIT\qquad $^{3}$UT Austin}}
\date{}

\begin{document}
\maketitle

\input{body_v2.tex}

\input{main.bbl}
\input{appendix_v2.tex}

\end{document}

%% file: body_v2.tex
\begin{abstract}
Policy gradient (\Pg{}) methods such as GRPO dominate LLM fine-tuning, yet recent work showed that Evolution
Strategies (\Es{}) also scale to fine-tuning models with billions of parameters~\citep{esatscale2025}. Which
method needs fewer samples, and on which tasks, remains unclear. To answer this question, we view the
two methods as estimators of the same gradient that differ in where they explore. Because \Es{} perturbs
\emph{parameters}, its noise naively grows with the number of parameters, and because \Pg{} samples
\emph{actions}, its noise grows with the number of generated tokens. Only noise along the directions that change
the reward slows learning. Once this is taken into account, the cost of \Es{} depends on the
\emph{effective dimension}, the number of independent skills a task requires rather than its parameter count,
while the cost of \Pg{} depends on the \emph{effective horizon}, which counts only the tokens that share both
credit and weights with the reward-carrying ones. To test these predictions, we
train both methods on a synthetic task in which the number of skills and the
effective horizon can be varied separately. \Pg{} is cheaper when the horizon is short, \Es{} becomes cheaper
once it is long enough, and this crossover moves to longer horizons as the number of skills grows. We then
measure the effective horizons of 20 NLP, mathematical reasoning, and agentic tasks and compare the two methods on
them with Qwen2.5-7B. \Pg{} leads on short-horizon tasks, the two tie on the moderate-horizon math ones, and no task has a horizon in the regime where the theory predicts an \Es{} advantage. Terminal-Bench has a far longer horizon. There, in a preliminary comparison with Qwen3.5-4B, \Es{} without any
credit assignment reaches a higher success rate than the \Pg{} checkpoint released with
Tmax~\citep{ivison2026tmax}, as the theory predicts. We conclude that \Es{} should have the advantage on
long-horizon tasks with few independent skills and a nonvanishing reward signal. Current benchmarks are only at the onset of this regime,
so \Es{} should become more competitive as agentic tasks grow longer.
\end{abstract}

\section{Introduction}\label{sec:intro}
Policy gradient (\Pg{}) methods such as GRPO dominate reinforcement-learning fine-tuning of language
models~\citep{shao2024grpo,deepseekr1}. Nonetheless, they are known to struggle with credit assignment, and
this is an especially prominent issue in long-horizon agentic tasks such as Terminal-Bench~\citep{merrill2026terminalbench}. Tmax reported that naive GRPO struggled to remain stable in long-horizon agentic training. It succeeded only
after masking tokens whose training and inference probabilities disagree, reducing that mismatch, and
enlarging the rollout groups~\citep{ivison2026tmax}. T1 relied on stabilized actor--critic training and rewards for partial task
completion~\citep{yang2026t1}, and TRACE assigned credit to individual tool actions~\citep{tao2026trace}. Taken
together, these works showed that training \Pg{} successfully on long-horizon tasks requires considerable additional
engineering. This difficulty is rooted in how \Pg{} assigns credit. Under terminal credit, every generated token
enters the \Pg{} gradient, so even tokens that do not affect the reward add variance to it.

\begin{figure}[t]\centering
\includegraphics{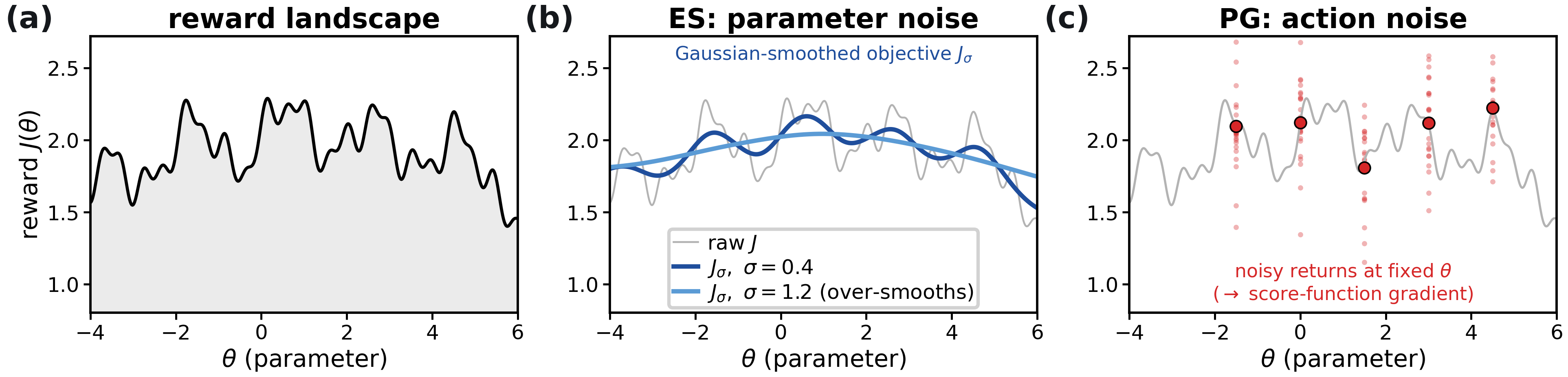}

\caption{\textbf{Parameter-space and action-space exploration.} \Es{} and \Pg{} estimate nearly the same gradient but draw
their noise from different spaces. \textbf{(a)} The expected reward landscape
$J(\theta)$. \textbf{(b)} \Es{} perturbs the parameters with Gaussian noise of scale $\sigma$ and estimates the
gradient of the smoothed objective $J_\sigma(\theta)=\mathbb E_{\varepsilon}[J(\theta+\sigma\varepsilon)]$. \textbf{(c)} \Pg{} samples actions and estimates the gradient of $J$. The \Es{} target gradient approaches the \Pg{} target gradient as $\sigma\to0$.}

\label{fig:concept}\end{figure}

Evolution Strategies (\Es{}) follow a different principle (Fig.~\ref{fig:concept}). \Es{} perturbs the model's parameters and scores each
perturbed model by its total return~\citep{salimans2017es,sehnke2010pgpe}, and therefore does not need to assign credit
to individual tokens. Recent work showed that \Es{} scales to LLM fine-tuning at the billion-parameter
level~\citep{esatscale2025}, but its performance relative to \Pg{} was mixed, and it is unclear which regime
favors \Es{} and which favors \Pg{}. Long horizons are a natural candidate for \Es{}. In principle, the horizon problem that affects \Pg{} is largely absent for \Es{}. When reward-irrelevant
actions leave the return of each perturbed model unchanged, they also leave the \Es{} estimate unchanged,
whereas under terminal credit the same actions still add noise to the \Pg{} estimate.

However, \Es{} pays a cost of its own. The two methods explore in different currencies. \Es{} explores in
\emph{parameter} space and pays for \emph{dimension}, while \Pg{} explores in \emph{action} space and pays for
\emph{horizon}~\citep{vemula2019contrasting}. For an LLM with billions of parameters, the dimension cost would
seem prohibitive. In practice, only the variance along reward-relevant directions affects learning,
and fine-tuning is known to have low intrinsic dimension~\citep{aghajanyan2021intrinsic}. Indeed,
\citet{blessingdim} found that the fine-tuning landscape is low-dimensional in curvature. A few
high-curvature directions dominate the reward improvement, so reward-improving perturbations remain easy to
find with small populations even for models with billions of parameters. We therefore
describe the two costs by an effective dimension $\Deff{}$, as proxied by the number of independent reward-carrying
decisions, and an effective horizon $\Teff{}$, as proxied by the number of token positions that contribute
reward-relevant \Pg{} variance. Unlike the sequence length, $\Teff{}$ counts a reward-irrelevant token only when
it receives credit from the reward and its gradient overlaps the reward-relevant directions. In an
autoregressive LLM this overlap is expected, because the same weights generate every token.

Because longer tasks also tend to be harder, the two costs are difficult to separate in real models. To study them
independently, we design a synthetic task with $r$ independent skills. In each rollout the policy is given one
skill and generates $k$ reward-carrying tokens and $F$ filler tokens. The number of skills, the filler
length, the parameter count, the credit scheme, and the overlap between filler and reward-carrying tokens can
each be varied on their own. We tune \Es{} and \Pg{} separately in every setting and measure the number of
rollouts each needs to solve the task to a threshold accuracy. Changing the parameter count leaves the cost of both methods unchanged. Adding skills raises the cost of
\Es{} much more than that of \Pg{}, while adding filler raises the cost of \Pg{} and leaves that of \Es{}
unchanged. As a result, \Pg{} is cheaper at short horizons and \Es{} at very long ones, and the crossover moves
to longer horizons when the task has many skills (Section~\ref{sec:phase}). \Es{} is therefore the better choice when
the effective horizon is long relative to the number of independent skills and the reward still carries a
signal.

Finally, we examine where real tasks fall. We measure the effective horizon of 20 NLP, mathematical reasoning, and
agentic tasks, each in the setup of its training run, and compare \Es{} and \Pg{} after tuning and training
both under the same setups. Current benchmarks sit at short to moderate effective horizons, where \Pg{} is expected
to be cheaper, and \Pg{} is ahead or tied on nearly all of them. The analysis predicts that an \Es{} advantage should appear on much longer tasks that still carry a learning signal, so we take a first step with Terminal-Bench, whose episodes run far longer and whose effective horizon we measure in the thousands of tokens. With Qwen3.5-4B, a base model strong enough
to obtain rewards, vanilla \Es{} reaches a higher success rate than the \Pg{} checkpoint released with
Tmax~\citep{ivison2026tmax} in a preliminary comparison, without any credit assignment~\citep{schweighofer2026es} (Section~\ref{sec:atlas}). These
results are consistent with the prediction from the testbed that the horizon is the axis along which \Es{} gains the
advantage.

\paragraph{Contributions.} In summary, this paper makes the following contributions.
\begin{itemize}\setlength\itemsep{1pt}
\item We cast \Es{} and \Pg{} as two estimators of the same gradient and show that \Es{} pays for the
effective dimension $\Deff{}$ and \Pg{} for the effective horizon $\Teff{}$. The effective horizon grows only
where reward-irrelevant tokens share both credit and weights with reward-carrying ones (Sections~\ref{sec:scalings}
and~\ref{sec:blessing}).
\item We build a multi-skill testbed in which the number of skills and the number of reward-irrelevant tokens
are set separately (Section~\ref{sec:task}). With both methods tuned in every setting, it shows that \Es{} is
cheaper beyond a crossover horizon that grows with the number of skills (Section~\ref{sec:phase}).
\item We measure $\Teff{}$ on 20 real tasks in the setups in which they are trained and compare tuned \Es{}
and \Pg{} on 14 of them at 7B. These horizons are short to moderate, and \Pg{} is ahead or tied on all 14, as expected.
The horizon estimate tracks the measured noise of \Pg{} updates, and a direct control shows that the comparison does not
depend on training low-rank adapters (Section~\ref{sec:atlas}).
\item We take a first step beyond these horizons on Terminal-Bench, where we measure an effective horizon in the thousands of tokens. In a preliminary comparison with Qwen3.5-4B~\citep{schweighofer2026es},
vanilla \Es{} reaches a higher success rate than the \Pg{} checkpoint released with Tmax, which is consistent with an
\Es{} advantage at long horizons (Section~\ref{sec:atlas}).
\end{itemize}
Together, these results place current benchmarks at the onset of the regime in which \Es{} gains the
advantage, and they give a first indication that this advantage can be unlocked at longer horizons. As agentic tasks grow longer, and provided they still yield a reward signal, \Es{} offers a promising way to
train long-horizon agents without precise credit assignment. \Es{} can thus expand the scope of LLM fine-tuning, and of
AI applications in general, to tasks whose horizons are too long for \Pg{} to train efficiently.

\section{Two exploration currencies: parameters and actions}\label{sec:scalings}
We fine-tune a policy $\pi_\theta$ with parameters $\theta\in\mathbb R^{D}$. Given an input, it generates a
sequence of $T$ tokens $a=(a_1,\dots,a_T)$ and receives a scalar reward $R(a)\in[0,1]$ at the end, as in
outcome-reward fine-tuning of an LLM. The objective is the expected reward
$J(\theta)=\mathbb E_{a\sim\pi_\theta}[R(a)]$, with gradient $g^\star=\nabla_\theta J(\theta)$. \Es{} and
\Pg{} both estimate $g^\star$ from sampled rollouts, and they differ in what they sample.

\subsection{Gradient estimates in the two exploration spaces}\label{sec:unify}
\paragraph{Evolution strategies.} \Es{}~\citep{salimans2017es} samples Gaussian perturbations of the
parameters, $\theta+\sigma\varepsilon$ with $\varepsilon\sim\mathcal N(0,I_D)$, and weights each perturbation
by the return of the perturbed model,
\begin{equation}\label{eq:es}
\hat g_{\Es}=\tfrac1\sigma\,R\big(a;\theta+\sigma\varepsilon\big)\,\varepsilon ,
\qquad
\mathbb E[\hat g_{\Es}]=\nabla_\theta J_\sigma ,
\end{equation}
where $J_\sigma(\theta)=\mathbb E_{\varepsilon}[J(\theta+\sigma\varepsilon)]$ is the Gaussian-smoothed objective.

\paragraph{Policy gradient.} \Pg{}~\citep{williams1992reinforce} holds the parameters fixed, samples a
trajectory $a\sim\pi_\theta$, and weights the gradient of each token's log probability by its advantage,
\begin{equation}\label{eq:pg}
\hat g_{\Pg}=\textstyle\sum_{t=1}^{T}\hat A_t\,\psi_t ,
\qquad
\mathbb E_{a\sim\pi_\theta}[\hat g_{\Pg}]=g^\star ,
\qquad \psi_t\equiv\nabla_\theta\log\pi_\theta(a_t\mid s_t).
\end{equation}
We call $\psi_t$ the \emph{score} of token $t$. With terminal credit every token receives the same advantage
$\hat A_t=R(a)-b$, where the baseline $b$ is the mean reward of the $G$ rollouts in the update. More localized
credit lowers the advantage of tokens that do not affect the reward.

\paragraph{Two exploration currencies (Fig.~\ref{fig:concept}).} \Es{} explores in parameter space and
\Pg{} in action space, so their noise grows with different quantities, as the next section shows. Their
targets also differ. \Es{} estimates the gradient of the smoothed surrogate $J_\sigma$, which approaches
$g^\star$ as $\sigma\to0$, while \Pg{} estimates $g^\star$ itself. In practice the two signals can differ,
because $J_\sigma$ smooths the reward landscape. To compare the methods by their noise, we assume throughout
the analysis that the two targets agree, so that the estimators differ only in their variance. Variance then
sets how many rollouts each method needs to reach a solution, while the exploration mechanism can decide which
solution it reaches. Appendix~\ref{app:dynamics} shows in training runs where the targets differ, on decisions where the policy
is confident in a wrong action.

\subsection{The naive dimension and horizon curses}\label{sec:theory}
Each estimator's noise naively grows with the size of the space it explores. Let $D$ be the number of
parameters, $T$ the number of generated tokens, and $\kappa$ the score energy of a single token, defined in
Eq.~\ref{eq:kappa} below. At a fixed $\theta$, with matched per-update sample counts ($P$ perturbations for
\Es{}, $G$ rollouts for \Pg{}; Table~\ref{tab:duality}) and under the covariance assumptions of
Appendix~\ref{app:tax}, the two variances scale as
\begin{equation}\label{eq:scalings}
\operatorname{tr}\mathrm{Cov}\big(\hat g_{\Es}\big)\ \propto\ D ,
\qquad
\operatorname{tr}\mathrm{Cov}\big(\hat g_{\Pg}\big)\ \propto\ T\cdot \kappa .
\end{equation}

\paragraph{The dimension curse of \Es{}.} Every \Es{} estimate is a scalar computed from the returns times
the perturbation, $\hat g_{\Es}=\Delta\,\varepsilon$. The scalar is $R(\theta+\sigma\varepsilon)/\sigma$ in
Eq.~\ref{eq:es} and $[R(\theta+\sigma\varepsilon)-R(\theta-\sigma\varepsilon)]/(2\sigma)$ for the
antithetic estimate we use in experiments. Either way the perturbation energy is $\mathbb
E\lVert\varepsilon\rVert^2=D$, so the variance grows linearly with the parameter count.

\paragraph{The horizon curse of \Pg{}.} The \Pg{} estimate sums $T$ scores, each weighted by the advantage,
so its variance scales as $\mathrm{Var}(\hat A)\,\kappa\,T$. Here $\kappa$ is the per-token \emph{score
energy},
\begin{equation}\label{eq:kappa}
\kappa\;\equiv\;\mathbb E_{a_t\sim\pi}\big\lVert u_t\big\rVert^2,\qquad
u_t\equiv\nabla_{z_t}\log\pi(a_t\mid z_t),
\end{equation}
the expected squared norm of the score with respect to the action distribution's own parameters $z_t$, such as
the logits of a categorical policy. The parameter-space score is $\psi_t=J_t^{\top}u_t$ with readout Jacobian
$J_t=\partial z_t/\partial\theta$, whose bounded magnitude we absorb into the proportionality constant.

\paragraph{The per-token cost depends on the action space.} Evaluating Eq.~\ref{eq:kappa} for categorical
logits, $u=\mathbf 1_a-p$, and for a Gaussian mean, $u=(a-\mu)/\sigma_a^2$, gives
\begin{align}
\text{categorical } (V\text{ symbols}):\ & \kappa=1-\textstyle\sum_v p_v^2\ \le\ 1
\quad(=1-1/V\text{ for uniform }\pi),\label{eq:Acat}\\
\text{continuous } (m\text{ dims}):\ & \kappa=m/\sigma_a^2\ \propto\ m.\label{eq:Acont}
\end{align}
The categorical score energy is bounded whatever the vocabulary size, while the Gaussian one grows with the
action dimension. For an LLM, the \Pg{} variance therefore grows with the number of tokens and not with the
size of the vocabulary.

\subsection{Multi-skill combination lock task as a testbed}\label{sec:task}
\begin{figure}[t]\centering
\includegraphics{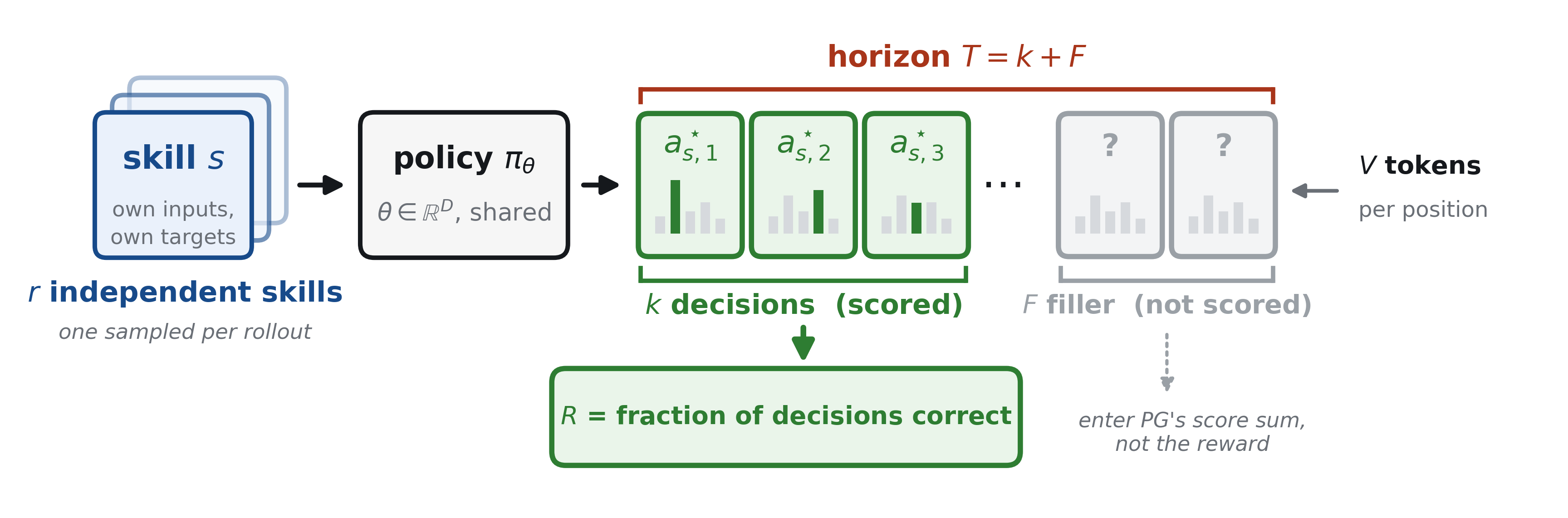}
\caption{\textbf{The controlled testbed.} The testbed sets the number of independent skills and the
number of reward-irrelevant tokens separately. The task has $r$ independent skills, each with its
own inputs and its own targets. A rollout samples one skill, and a policy with shared parameters
$\theta\in\mathbb R^{D}$ emits a categorical distribution over $V$ tokens at each of $T=k+F$ positions. The $k$
\emph{decision} positions have fixed targets, and the reward is the fraction of them that are correct. The $F$
\emph{filler} positions do not affect the reward but still enter the \Pg{} gradient. The controls are the
number of skills $r$, the decisions per skill $k$, the filler length $F$, the parameter count $D$, and the
vocabulary size $V$.}\label{fig:task}\end{figure}

To vary the quantities in these scalings one at a time, we design a \emph{multi-skill combination lock}
(Fig.~\ref{fig:task}). The task consists of learning $r$ skills, each of which is a combination to memorize. Each
combination has $k$ decision positions with fixed target tokens, and $F$ filler positions. A rollout samples one
skill uniformly, and the policy generates $T=k+F$ tokens from a vocabulary of size $V$. The reward is the
fraction of the skill's $k$ decisions that match their targets. The filler tokens do not affect the reward but still enter the \Pg{} gradient, like the tokens an LLM
generates around a final answer. The reward is given only at the end, as in outcome-reward fine-tuning. Skills are
independent, in the sense that their inputs are unrelated and their targets are drawn separately, so learning
one skill gives no information about another. The task therefore separates three quantities that a single
sequence confounds. The number of skills $r$ adds decisions that the update must set without lengthening any
rollout. The filler length $F$ lengthens every rollout without adding decisions. The number of decisions per
skill $k$ does both, and we use it as a control. We fix the output length so that the horizon can be set
directly. An LLM chooses its own length, so our benchmark probes (Section~\ref{sec:atlas}) use each example's actual
generated length.

The policy is a \emph{shared MLP} that maps a fixed
input code of each position to its $V$ logits, so all positions and skills share the same weights, as the
tokens of an LLM do. Decision codes are random. We control how much filler overlaps the decisions through the
cosine similarity between filler and decision codes. \emph{Local} filler codes are similar to the decision codes of their own skill, as generated text resembles its prompt, and \emph{global} filler codes are random
directions shared by all skills. Appendix~\ref{app:robust} gives the input codes, the network, and how both
filler types are constructed.

\paragraph{Experimental setup.} Both methods use $64$ rollouts of one sampled skill per update. \Es{} uses $32$
antithetic pairs with rank-shaped returns and scores each perturbed policy on one sampled trajectory. \Pg{}
uses REINFORCE with a mean baseline and an entropy bonus. In every setting we tune the step size and perturbation scale of \Es{} and the step size and entropy bonus
of \Pg{} on held-out seeds, then evaluate the selected configuration on fresh seeds.
We report sample cost, the number of rollouts until the expected reward first reaches 90\%. The expected
reward is the mean probability the policy assigns to the targets over all $r\cdot k$ decisions, and we check
it after every update. When the number of skills varies we divide by $r$ and report the cost per skill. We use sample cost rather than accuracy at a fixed budget because both tuned methods solve every setting and
differ only in how many rollouts they need. Sections~\ref{sec:blessing} and~\ref{sec:phase} report the results
under this setup. Appendix~\ref{app:robust} lists the hyperparameter grids, the numbers of seeds, and the
stopping rule for tuning, together with controls that vary the number of rollouts per update, the number of
decisions per skill, and the success criterion.

\begin{figure}[t]\centering
\includegraphics{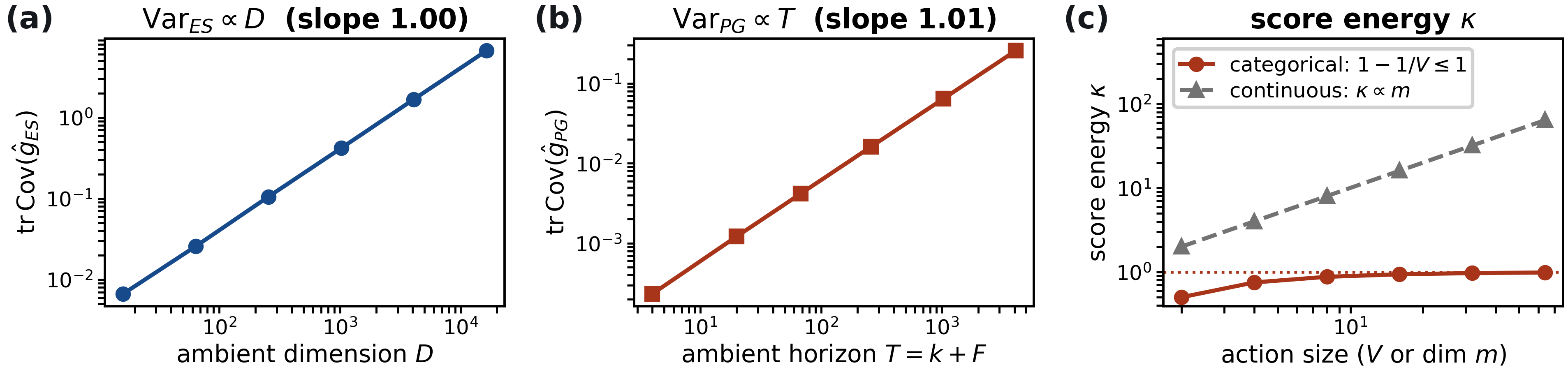}

\caption{\textbf{Variance at a fixed policy.} Counted over all
directions, the noise of \Es{} grows with the parameter count and that of \Pg{} with the sequence length. Panels (a) and (b)
use the linear lift at initialization with one skill, $k=4$, and $V=5$, and measure $\operatorname{tr}\mathrm{Cov}(\hat g)$ without optimization. \textbf{(a)} \Es{} variance grows linearly
with $D$ over a $1000\times$ range. \textbf{(b)} \Pg{} variance grows linearly with $T$.
\textbf{(c)} Per-step score energy $\kappa$ is bounded for categorical actions and grows linearly with
continuous action dimension.}

\label{fig:scalings}\end{figure}

\paragraph{Validating the naive scalings.}
Before training, we check that the testbed reproduces the naive scalings of Eq.~\ref{eq:scalings} at a
fixed policy. Testing the dimension curse requires changing the parameter count while keeping the task fixed. Widening the MLP
changes the policy and its gradients along with $D$, so for this check we use a \emph{linear lift},
$\mathrm{logits}_t=A_t\theta$, where each position has its own fixed random matrix $A_t\in\mathbb R^{V\times D}$.
At $\theta=0$ the policy is uniform for every $D$, and the reward-relevant subspace keeps dimension $(V-1)k$ per
skill, so only $D$ changes. The true gradient is also available in closed form. We use one skill, $k=4$ targets and $V=5$. For \Es{} we vary $D$ from $16$ to
$16{,}384$ with no filler. For \Pg{}, we fix $D=256$ and vary $F$ from $0$ to $4096$. As Eq.~\ref{eq:scalings} predicts, the variance of \Es{} grows linearly with $D$ and that of \Pg{} linearly
with $T$ (Fig.~\ref{fig:scalings}). Panel (c) shows that the categorical score energy approaches $1$ while the continuous one grows
linearly with the action dimension, as Eqs.~\ref{eq:Acat}--\ref{eq:Acont} predict. All three panels measure total
variances. With the testbed defined, the next section defines the effective dimension and the effective
horizon on it, which count only the part of these variances that affects learning.

\section{Effective dimension and effective horizon}\label{sec:blessing}

\begin{figure}[tbp]\centering
\includegraphics{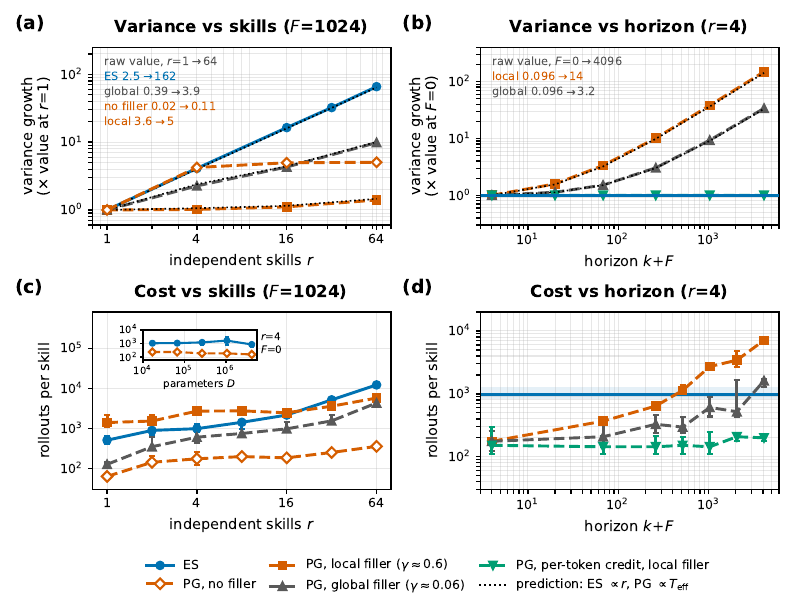}
\caption{\textbf{What each method pays for.} \Es{} pays mainly for the number of skills and \Pg{} for the effective
horizon, and neither pays for the parameter count. Left, the number of independent skills $r$; right, the
horizon. Top, reward-relevant variance at initialization, relative to its value at $r=1$ (a) or $F=0$ (b);
bottom, trained cost in rollouts per skill to reach expected reward $0.9$. \Pg{} uses terminal credit unless marked per-token. \textbf{(a)} Variance against $r$
at $F=1024$. Dotted lines are $\propto r$ for \Es{} and, for \Pg{}, the measured
no-filler variance plus the filler term $\gamma\beta F$ of $\Teff{}$ (Section~\ref{sec:blessing}), with the overlap
$\gamma$ measured at each $r$ and no fitted parameter. \textbf{(b)} Variance against $k+F$ at $r=4$, with
$\Teff{}=k+\gamma\beta F$ dotted. \textbf{(c)} Cost per skill against $r$ at $F=1024$; inset, against the parameter count $D$ at $r=4$
without filler. \textbf{(d)} Cost per skill against the horizon at $r=4$. Error bars are interquartile ranges
over $8$ (c) and $16$ (d) seeds.}

\label{fig:blessing}\end{figure}

Read naively, Eq.~\ref{eq:scalings} says that \Es{} cannot
scale to large models. However, only part of each variance affects actual learning. Noise along a direction in which the reward does not change moves the parameters but not the reward, so only noise inside the \emph{reward-relevant subspace} $S$ competes with the signal. Here we
take $S$ to be the row space of the Jacobian of the decision logits with respect to $\theta$, over the
decision positions of all $r$ skills. The subspace $S$ contains every direction along which changing $\theta$ changes the probability of some rewarded decision.
Equivalently, $S$ is the span of all possible decision scores and has dimension $(V-1)rk$. We project each estimate onto $S$ with the orthogonal projection $P_S$. The projected variance of \Es{} then scales with $\dim S$ rather than $D$. We therefore define the effective
dimension $\Deff{}$ as the number of independent reward-carrying decisions, $\dim S/(V-1)$. The projected variance of \Pg{} sums, over token positions, the
score energy each one places inside $S$. A reward-carrying token contributes in full, and a reward-irrelevant
token contributes only the fraction of its score energy that falls inside $S$, weighted by the credit it
receives. We call this sum the effective horizon $\Teff{}$. Appendix~\ref{app:tax} derives $\Teff{}$ from three assumptions, which the next paragraph states.

The derivation does not depend on the architecture and requires only that tokens are sampled independently
given the parameters, that the reward does not depend on the reward-irrelevant tokens, and that the credit of
such a token does not depend on the token itself. The result therefore applies to the shared MLP, on which we test
it.\footnote{In the linear lift the Jacobian does not depend on $\theta$, so $S$ is fixed. In the shared MLP it
does, and we compute $S$ at initialization, where all variances in this section are measured, so the \Es{}
argument holds to first order in $\sigma$. We use the MLP rather than the lift because in the lift every
position has its own random readout, so a filler score overlaps $S$ only by chance, with $\gamma\approx\dim
S/D$ as derived in Appendix~\ref{app:tax}, and the weight sharing on which $\Teff{}$ depends cannot be varied. The MLP is also
the network we train, so its variances can be compared with the trained costs.}

\paragraph{Effective dimension for \Es{}.}
Suppose reward depends on the perturbation $\varepsilon$ only through its projection onto a subspace $S$.
The total and projected variances of the antithetic estimate then scale as

\begin{equation}\label{eq:esrel}
\operatorname{tr}\mathrm{Cov}\big(\hat g_{\Es}\big)\ \propto\ \mathbb E[\Delta^2]\,D ,
\qquad
\underbrace{\operatorname{tr}\mathrm{Cov}\big(P_{S}\,\hat g_{\Es}\big)}_{\text{reward-relevant}}\ \propto\
\mathbb E[\Delta^2]\,\dim S ,
\qquad \mathbb E[\Delta^2]\le\tfrac1{4\sigma^2}\ (R\!\in\![0,1]).
\end{equation}
The bound on the perturbation signal $\mathbb E[\Delta^2]$ follows from a reward in $[0,1]$. Increasing
$D$ at a fixed task therefore adds variance only outside $S$, where it does not change the expected update,
whereas adding skills enlarges $S$ itself. We check both effects on the shared MLP. We widen the hidden layer and shrink $\sigma$ as one over the square root of the width, so that one
perturbation moves the policy equally at every width. The total variance then grows in proportion to $D$, as on
the linear lift (Fig.~\ref{fig:scalings}a), while the reward-relevant variance stays roughly constant. Figure~\ref{fig:blessing}a shows how the reward-relevant variance scales with the number of skills.
Each curve is divided by its own value at $r=1$, so the panel shows how fast each method's variance grows.
The absolute levels depend on each method's step size and perturbation scale, so we compare them through the
trained cost in Fig.~\ref{fig:blessing}c. The variance of \Es{} grows in proportion to $r$.

The trained cost of \Es{} confirms both predictions. The cost does not change with the parameter count
(Fig.~\ref{fig:blessing}c, inset), in line with \citet{blessingdim}, but it rises with the number of skills
(Fig.~\ref{fig:blessing}c). The reward of each rollout depends only on the decisions of the sampled skill, yet the \Es{} perturbation
also moves the decisions of every other skill. Those components enter the estimate as noise. We therefore
identify $\Deff{}$ with the number of independent reward-carrying decisions, $r\cdot k$ in the testbed.
Appendix~\ref{app:robust} repeats part of the grid with $k=16$ and confirms that the cost of \Es{} depends on
this product. We vary it through $r$, because changing $k$ would also lengthen every rollout.

\paragraph{Effective horizon for \Pg{}.}
For \Pg{}, we apply the same projection to $\hat g_{\Pg}=\sum_t\hat A_t\psi_t$.
Under the assumptions in Appendix~\ref{app:tax}, the variance contributions give

\begin{equation}\label{eq:pgdual}
\operatorname{tr}\mathrm{Cov}\big(\hat g_{\Pg}\big)\ \propto\ \mathrm{Var}(\hat A)\,\kappa\,T ,
\qquad
\underbrace{\operatorname{tr}\mathrm{Cov}\big(P_{S}\,\hat g_{\Pg}\big)}_{\text{reward-relevant}}\ \propto\
\mathrm{Var}(\hat A)\,\kappa\,\Teff{} ,\qquad \mathrm{Var}(\hat A)\le\tfrac14 .
\end{equation}
The $k$ reward-carrying positions of the visited skill contribute in full. A filler position contributes no mean gradient under the
reward-independence assumption, but can still contribute variance within $S$. We describe its contribution
with two factors, $\gamma$, the fraction of its score energy that lies in $S$, and $\beta$, the relative
credit assigned to it. Weight sharing controls the first factor and credit assignment the second.

In the uniform case, these factors give $\Teff{}=k+\gamma\beta F$. In general, each filler position has its own factors, and
$\Teff{}=k+\sum_{\text{filler}}\beta_t\gamma_t\kappa_t$,
as derived in Appendix~\ref{app:tax}. Filler therefore increases the effective horizon only when it both receives credit
and contributes score energy along reward-relevant directions (Table~\ref{tab:factors}).

\begin{table}[t]
\centering\small
\caption{\textbf{When \Pg{} pays for filler.} Filler lengthens the effective horizon only when it both
receives credit ($\beta>0$) and overlaps the reward-relevant directions through shared weights ($\gamma>0$).}
\label{tab:factors}
\begin{tabular}{r|cc}
& $\gamma\!\to\!0$ (no overlap) & $\gamma\!>\!0$ (shared weights)\\\hline
$\beta\!\to\!0$ (per-token credit) & free & free\\
$\beta\!\approx\!1$ (terminal credit) & free & \textbf{paid for}\\
\end{tabular}
\end{table}

\paragraph{How the number of skills enters each count.}
Adding skills enlarges $S$ for both methods, but they pay for it differently. The \Es{} perturbation is isotropic and fills every direction of $S$, so its count $\dim S$ grows in
proportion to $r$. The decision
tokens of \Pg{} place their noise in the directions of the skill that produced them, so they contribute $k$
regardless of the number of skills. A filler token reaches the directions of other skills only to the extent that
its score overlaps them. Because $\gamma$ is measured against the $S$ of all skills, it grows with $r$, giving
$\Teff{}(r)=k+\gamma(r)\beta F$. \Pg{} therefore pays for skills only through the part of $S$ that its filler
reaches. A second, smaller dependence comes from sampling one skill per update. The gradient of the sampled
skill differs from the average over all skills, which adds variance once $r>1$. This sampling term grows little
thereafter and changes neither count.

\paragraph{Testing the sharing and credit factors.}
This view predicts that the reward-relevant variance of \Pg{} grows with filler as $k+\gamma\beta F$. We measure this variance and the overlap $\gamma$ at initialization on the shared MLP with four skills. The
prediction matches the measured variance in every condition with no fitted parameter
(Fig.~\ref{fig:blessing}b, Table~\ref{tab:gates}). Filler inflates the variance most when it receives terminal
credit and overlaps the decisions strongly, less when its overlap is small, and not at all under per-token
credit. Either factor alone is therefore enough to make filler free. For the dependence on skills (Fig.~\ref{fig:blessing}a), we add the filler term $\gamma(r)\beta F$ to the
measured no-filler variance, which already contains the term from sampling one skill per update. The prediction tracks the measured growth because the overlap with the $S$ of all
skills grows with $r$, strongly for global filler and little for local filler.

The trained cost of \Pg{} responds to the two factors in the same way (Fig.~\ref{fig:blessing}d). With terminal credit, the cost rises with the filler length until it exceeds that of \Es{}. The crossing comes sooner
for local filler than for global filler, whose overlap is smaller. With per-token credit it does not rise at
all. The cost is therefore set by the effective horizon rather than the sequence length. A short generation
whose filler overlaps the decisions can cost \Pg{} as much as a much longer one whose filler does not. Because the effective horizon sets the cost, we
can place real tasks on a single horizon axis in Section~\ref{sec:atlas}. The same reasoning implies that better credit
assignment removes \Pg{}'s horizon cost, whereas shorter generations need not. Appendix~\ref{app:pgvariants} compares REINFORCE with RLOO, GRPO, and Dr.GRPO and finds the same effective
horizon for all four, although their total variances differ widely.

\paragraph{Putting the two costs side by side.}
Equations~\ref{eq:esrel} and~\ref{eq:pgdual} have the same form. Each method's reward-relevant variance is
a bounded noise term times a count. Because the noise terms are bounded, how each method's cost scales with
task structure is set by its count, up to method-specific factors.
\begin{equation}\label{eq:llm}
\text{\Es{} cost grows with}\ \ \Deff{}=r\cdot k ,\qquad \text{\Pg{} cost grows with}\ \ \Teff{}=k+\gamma(r)\beta F .
\end{equation}
Two axes separate the methods. \Es{} pays in full for the number of independent skills, which \Pg{} pays for
only through $\gamma(r)$, and only \Pg{} pays for filler. Because \Es{} pays more for skills and \Pg{} for
filler, neither method should be cheaper in every regime. Section~\ref{sec:phase} trains both methods over a grid of
skill counts and filler lengths to find where the two costs cross and how the crossing moves with the number
of skills. Real tasks are harder to place, because their directions and tokens matter to different degrees
rather than all or nothing. Section~\ref{sec:atlas} therefore measures their horizon count with a probe and
reports how far their dimension count can be measured.

\section{Trained sample cost across effective dimension and horizon}\label{sec:phase}

\begin{figure}[t]\centering
\begin{minipage}[t]{0.5405\textwidth}\vspace{0pt}\centering
\includegraphics{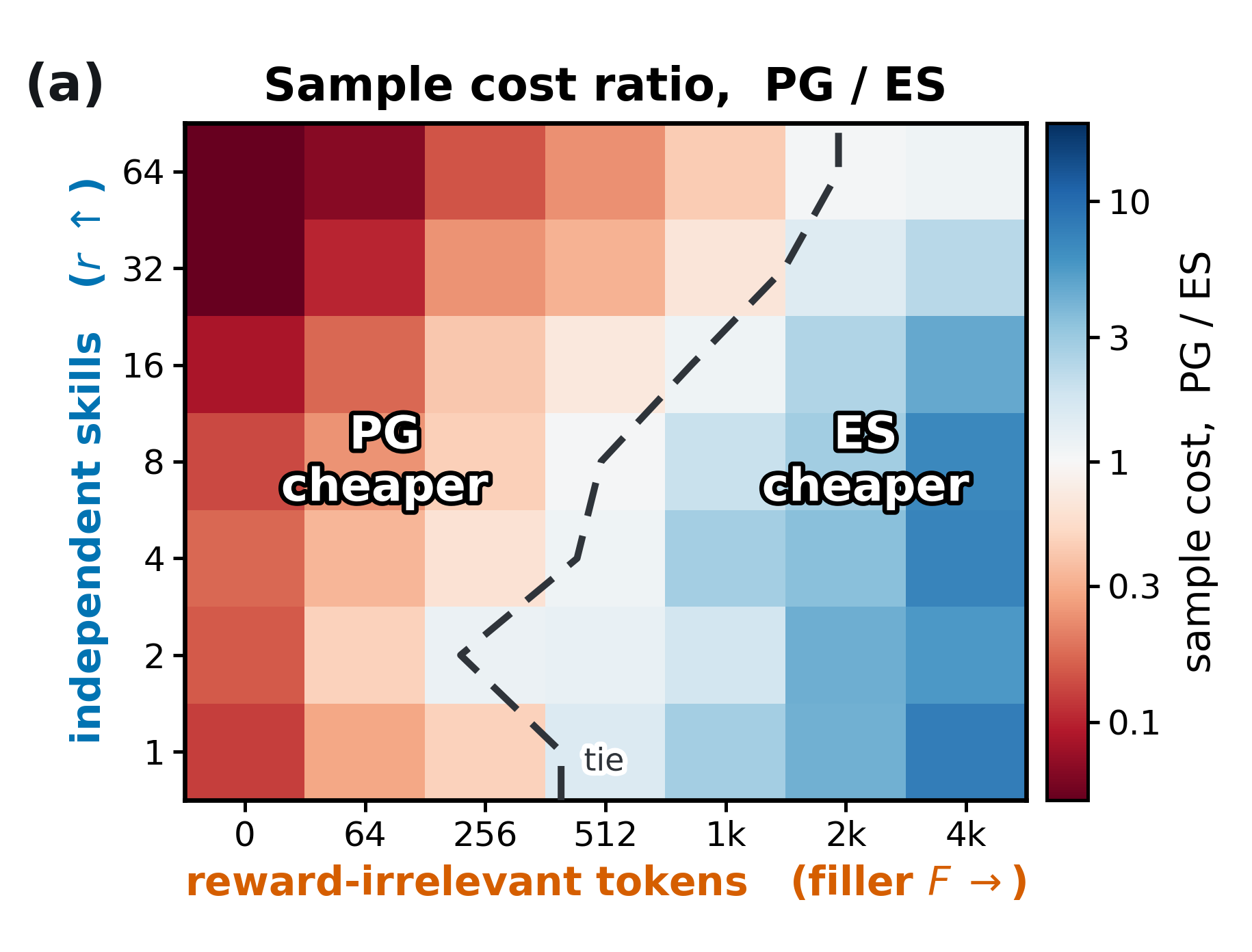}
\end{minipage}\hspace{0.005\textwidth}%
\begin{minipage}[t]{0.4535\textwidth}\vspace{0pt}\centering
\includegraphics{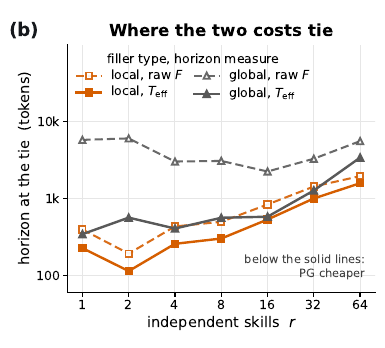}
\end{minipage}
\caption{\textbf{Trained sample cost on the shared MLP.} \Es{} becomes cheaper than \Pg{} beyond a crossover
horizon, which grows with the number of skills and is set by the effective horizon rather than the number of
filler tokens. All runs use $k=4$ and terminal credit, with both methods tuned in every cell. Sample cost is the number of
rollouts until the expected reward of the policy reaches $0.9$.
\textbf{(a)} Ratio of the costs of \Pg{} and \Es{} across the number of independent skills and the filler
length, with local filler. The dashed line marks where the two methods tie. \textbf{(b)} The horizon at which
the two methods tie, against the number of skills, for local and global filler. In filler tokens (dashed)
the two filler types differ by up to an order of magnitude. In effective tokens,
$\Teff{}=k+\gamma F$ with the overlap $\gamma$ measured against the reward-relevant directions of all skills
(solid), they agree within about a factor of two at every number of skills except two, and the tie horizon
grows with the number of skills. Table~\ref{tab:grid} lists the costs.}

\label{fig:plane}\end{figure}

Section~\ref{sec:blessing} measured each cost along one axis at a time. To see where the two costs cross, we
now train both methods on the shared MLP over a grid of skill counts and filler lengths, with $k=4$
decisions per skill\footnote{The
decisions per skill could serve as a third axis. Appendix~\ref{app:robust} repeats part of the grid with $k=16$
and finds that the cost of \Es{} depends on $r\cdot k$ and the tie on the effective horizon, within a factor
of about two.} and a single reward at the end of each rollout. With $k$ fixed, the number of skills sets the
effective dimension $\Deff{}=r\cdot k$, and the filler length sets the effective horizon. Figure~\ref{fig:plane}a shows the ratio of
the two sample costs with local filler.

Each row of the plane follows the pattern that Section~\ref{sec:blessing} predicts. Without filler \Pg{} is cheaper. As filler is added, its cost climbs while that of \Es{} stays flat, until
the two cross at the dashed line. Moving up the plane, the crossing stays at a similar horizon while skills are
few and moves to longer horizons only when they are many. With many skills, \Es{} has many more independent
decisions to set, so it needs a longer horizon before its insensitivity to filler outweighs that cost. The
shift is modest because \Pg{} also pays for skills. All skills share one network, so an update for one skill
can push the decisions of other skills onto wrong tokens, and tuning keeps the step of \Pg{} smaller as skills
are added. Appendix~\ref{app:dynamics} traces these locks in training runs and shows how the tuned step of
\Pg{} shrinks with the number of skills. \Pg{} still pays less for skills than \Es{} does, which is also why its lead without filler widens only
once skills are many. A task with many more skills therefore needs only a modestly longer horizon before \Es{}
gains the advantage.

How far the crossing moves depends on where the filler noise falls. Local filler stays within the directions of the visited skill, whereas global filler overlaps the directions
of every skill. With global filler, the cost of \Pg{} per skill therefore grows with $r$ about as fast as that
of \Es{}, and the two methods stay close at every number of skills (Fig.~\ref{fig:plane-global}). The two filler
types cross at very different filler lengths but at similar effective horizons $k+\gamma F$
(Fig.~\ref{fig:plane}b). The tie is therefore located by $\Teff{}$ rather than by the number of tokens. The distinction matters for real
tasks, because on the math, puzzle, and agentic tasks of Section~\ref{sec:atlas} most of the filler noise lies in
directions shared with other examples, as with global filler.

\section{Where real fine-tuning tasks fall}\label{sec:atlas}\label{sec:levers}
The testbed identifies three conditions under which \Es{} should gain on \Pg{} in LLM fine-tuning.
(i)~\emph{Low dimension.} The update the task needs involves few independent skills, so that the dimension
cost of \Es{} stays small. A strong pretrained model makes this likely, since fine-tuning updates are known
to be low-dimensional~\citep{li2018intrinsic,aghajanyan2021intrinsic,hu2022lora,blessingdim}. (ii)~\emph{Long horizon.}
The reward arrives only at the end of a generation whose effective horizon lies beyond the crossover, of
several hundred to several thousand tokens, so that the horizon cost of \Pg{} is large. (iii)~\emph{A
learning signal.} The reward varies with the policy's behavior, so that either method can learn at all.

To see how far current benchmarks meet these conditions, we measure the effective horizon of 20 tasks,
examine what can be measured about their effective dimension, and compare the outcomes of training both
methods. The tasks cover language understanding and generation, on which we train
Qwen2.5-7B-Instruct~\citep{qwen2025qwen25} ourselves (ARC~\citep{clark2018arc},
CommonsenseQA~\citep{talmor2019csqa}, HellaSwag~\citep{zellers2019hellaswag},
BoolQ~\citep{clark2019boolq}, SQuAD~\citep{rajpurkar2016squad}, WebQuestions~\citep{berant2013webq},
XSum~\citep{narayan2018xsum}, AG News~\citep{zhang2015agnews}, WinoGrande~\citep{sakaguchi2020winogrande},
OPUS-100 translation~\citep{zhang2020opus100}, TriviaQA~\citep{joshi2017triviaqa}, and
IMDB~\citep{maas2011imdb}), two agentic environments, on which we train the same model
(ALFWorld~\citep{shridhar2021alfworld} and WebShop~\citep{yao2022webshop}), and the tasks of prior \Es{} work~\citep{esatscale2025}
whose outcomes it reported for both methods. The latter are Countdown~\citep{gandhi2024sos} with
Qwen2.5-7B-Instruct and mathematical reasoning with Qwen2.5-Math-7B (MATH-500~\citep{lightman2023verify},
Minerva Math~\citep{lewkowycz2022minerva}, OlympiadBench~\citep{he2024olympiadbench}, AIME 2024, and AMC 2023).

\begin{figure}[!t]\centering
\includegraphics{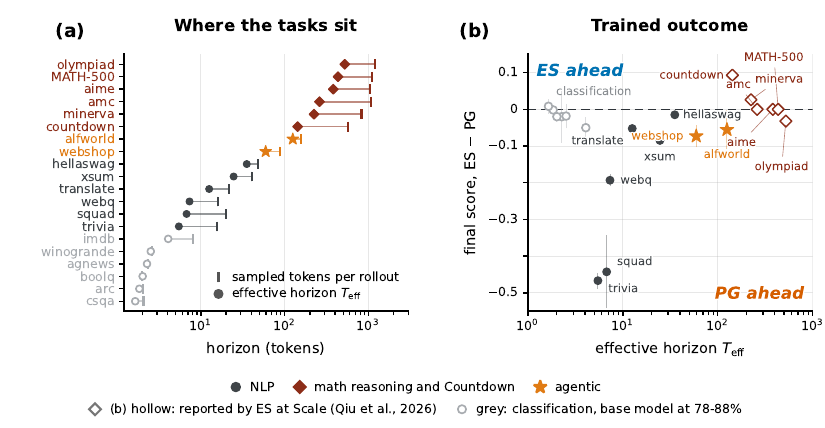}
\caption{\textbf{Real tasks on the horizon axis, and trained outcomes.} Current benchmarks sit at short to moderate effective
horizons, below the crossover, where \Pg{} should need fewer samples.
\textbf{(a)} Sampled tokens per rollout (tick) and effective horizon $\Teff{}$ (marker) of each task, measured in
the setup of its training run and ordered by $\Teff{}$ (values in Table~\ref{tab:atlasfinal}). Grey marks classification tasks, whose base accuracy is
$78\%$ to $88\%$.
\textbf{(b)} Final score of \Es{} minus that of \Pg{}. Filled: our comparisons at equal episodes, each method tuned (mean of three seeds, bars combine the two standard deviations).
Hollow: outcomes reported by \citet{esatscale2025}, whose math baselines are released RL checkpoints, not matched runs.}

\label{fig:measured}\end{figure}

\paragraph{Measuring the horizon.}
\Pg{}'s horizon cost comes from sampled tokens whose score vectors overlap the directions that change the
reward (Section~\ref{sec:blessing}). We approximate these directions by the span $S$ of the log-likelihood
gradients of the reference answers. Tokens sampled at the positions of a reference answer play the role of the decision tokens. Each generated
token then counts by how much of its score energy lies in $S$, relative to theirs (Appendix~\ref{app:probedef}). On the testbed, where $\Teff{}$ is known, this estimate matches it closely. We measure every task in the setup of its training run. Because the estimate depends on the reference
answers, we use it to order tasks rather than to place a boundary. On real models the estimate tracks the noise that $\Teff{}$ is meant to describe. Across the 20 tasks, its rank correlation with the measured noise-to-signal ratio of a \Pg{} update on the training objective is $+0.83$, and $+0.55$ within the 14 tasks on which we also train both methods (Appendix~\ref{app:pgnoise}).

\paragraph{Measuring the dimension.}
The number of directions spanned by the reference gradients equals the number of skills on the synthetic task, but on real tasks it instead reflects the style mismatch between model outputs and human-written references. What a skill count has to measure is how the reward
varies, so we also score the model's own rollouts, which finds about a dozen to several hundred skills on short-answer tasks, about $20$ on WebShop and about $130$ on ALFWorld
(Table~\ref{tab:skills}). On long reasoning tasks this needs about a thousand rollouts per problem, and on tasks
that the base model almost never solves there is no reward gradient to count. The tasks that come closest to the horizon condition are therefore exactly those on which the dimension
condition cannot be checked, so whether long agentic tasks with reasoning involve few independent skills remains open. Nevertheless, on every agentic task tested so far, \Es{} makes progress with a population of at most $32$, which suggests that these tasks do not suffer from a curse of dimensionality.

\paragraph{Where the tasks fall.}
Figure~\ref{fig:measured}a places the tasks on the horizon axis. Classification and multiple-choice tasks have
the shortest horizons, of a few tokens, and the other language tasks stay below their generation limits of a
few dozen tokens. Agents that emit a short action at each step reach up to about a hundred tokens, and Countdown and the math
tasks reach a few hundred. Together with the head-to-head comparisons below, this
indicates that none of these tasks reaches beyond the lower end of the crossover range. The gap between tick and marker shows how much of a rollout's noise lies outside the
reward-relevant directions. The gap is small for the action-only agents and larger for the math tasks, whose long
solutions contain many tokens that do not affect the reward. Terminal-Bench episodes are far longer than any of these rollouts, and we turn to them later in this section.

\paragraph{Training comparisons at equal samples.}
We train both methods on $14$ of the tasks with Qwen2.5-7B-Instruct for the same number of episodes and tune the step size or learning rate of each
(Fig.~\ref{fig:measured}b). Appendix~\ref{app:trained} gives the setup, the tuning ranges, the training length, and the score on
every task. On the language tasks \Pg{} trains low-rank adapters while \Es{} moves all weights, so we repeated all 12 of them with \Pg{} training all weights. Its score changes by at most $0.03$ on every task except AG News, where it improves, so the comparison does not depend on the adapters (Table~\ref{tab:fullpg}). At equal samples \Pg{} is ahead or tied on every task, as the analysis predicts below the crossover. Its leads are largest on short free-text tasks, where it quickly learns the answer format, and remain clear on the two agentic tasks. Its lead is larger on ALFWorld than on WebShop despite ALFWorld's longer horizon, as ALFWorld needs several times as many skills (Appendix~\ref{app:trained}). Three caveats limit this reading. First, these comparisons measure sample efficiency at a fixed budget, not the best score each method can attain. With a population of $30$, \Es{} matches the final score of \Pg{} on SQuAD at about ten times its samples (single seed, best of nine settings), so the large leads at short horizons reflect cost rather than a ceiling. Second, the horizon is confounded with the kind of task, and the size of \Pg{}'s lead does not follow the horizon reliably across these tasks. Third, \Es{} does not pull ahead on any of them, so they test only the regime in which \Pg{} is expected to lead. On math reasoning, prior work reports \Es{} close to released RL checkpoints~\citep{esatscale2025}, at unequal budgets.

\begin{figure}[t]\centering
\includegraphics{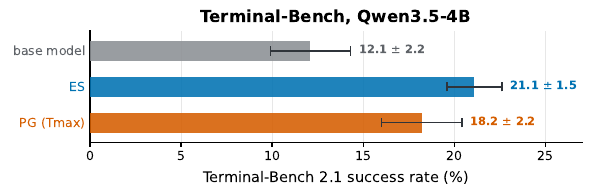}
\caption{\textbf{Performance comparison on Terminal-Bench 2.1.} Vanilla \Es{} improves on the base model at least as much as
the best released \Pg{} checkpoint for this model on Terminal-Bench 2.1, a task far longer than those of Fig.~\ref{fig:measured}. Results show the Qwen3.5-4B base model before training, after
\Es{} fine-tuning, and after \Pg{} training with the Tmax recipe~\citep{ivison2026tmax}. Each checkpoint is evaluated five times, and bars show the mean and the spread over the five evaluations. Results from \citet{schweighofer2026es}.}
\label{fig:tbench}\end{figure}

\paragraph{Terminal-Bench as a first test at a longer horizon.}
Terminal-Bench~\citep{merrill2026terminalbench} is much longer than the tasks discussed so far, but Qwen2.5-7B-Instruct solves
effectively none of its tasks, which leaves neither \Es{} nor \Pg{} with a learning signal. We therefore switch to
Qwen3.5-4B~\citep{qwen3.5}, which solves some of them, and report the comparison of \citet{schweighofer2026es} between \Es{} and Tmax~\citep{ivison2026tmax}, the open recipe for training terminal agents with \Pg{}. The base model samples about $18{,}000$ tokens per episode, more than ten times the longest rollout in Fig.~\ref{fig:measured}a. On the base-model rollouts released with Tmax, we measure an effective horizon of about $7{,}300$, some sixty times that of ALFWorld (Appendix~\ref{app:tbench}). The terminal's responses, which the model reads but does not sample, add further tokens to each episode.
Vanilla \Es{}, which uses no credit assignment, raises the success rate of the base model further than the \Pg{} checkpoint released with Tmax for the same
model (Fig.~\ref{fig:tbench}). Appendix~\ref{app:tbench} provides details on the setup.

The \Es{} run uses the implementation of \citet{esatscale2025} and none of the techniques commonly used to improve \Pg{} pipelines, such as active sampling, data filtering, asynchronous updates, or output length penalties.
The comparison concerns attainable performance rather than sample cost, because the budgets of the two pipelines are hard to put on equal terms. Active sampling and filtering change how many rollouts each \Pg{} update consumes and which tasks they come from, and an \Es{} rollout needs only a forward pass, whereas a \Pg{} update also needs backward passes. Neither rollout counts nor compute therefore give a fair common measure, and Appendix~\ref{app:tbench} lists the settings of both runs. The ranges of the two checkpoints overlap, and each method contributes a single checkpoint.
The \Pg{} baseline\footnote{\url{https://huggingface.co/allenai/tmax-4b}} comes from the original version of the Tmax recipe~\citep{ivison2026tmax}.
Even so, the ordering of the two methods changes. \Pg{} is ahead or tied on nearly every shorter task, whereas on this far longer task \Es{} finishes ahead on average without any of these techniques. This shift is in the direction the analysis predicts for a long horizon with a learning signal.

\paragraph{Takeaways.}
The results speak to each of the three conditions. (i)~\emph{Low dimension.} Prior work fine-tuned all
weights of a 7B model with a small \Es{} population and matched RL on math and Countdown~\citep{esatscale2025}.
Since the cost of \Es{} grows with $\Deff{}$, these reasoning tasks must need an update with few independent
skills, even though we cannot count them directly. (ii)~\emph{Long horizon.} Current benchmarks lie at or
below the lower end of the crossover, where \Pg{} should lead or tie, and it does on nearly all of them. A task is placed by its effective
horizon rather than its length, since much of a long rollout can lie outside the reward-relevant directions. (iii)~\emph{A learning signal.} A 7B base model almost never solves Terminal-Bench, so on long agentic tasks
the first obstacle for either method is obtaining a reward at all, before the horizon matters. Together, these results place current benchmarks at the onset of the regime in
which \Es{} gains the advantage. They point to long tasks with outcome rewards that a stronger base model
sometimes obtains as the place to look for it, and Terminal-Bench with Qwen3.5-4B, a first such test at a horizon in the thousands, already leans toward \Es{}.

\section{Related work}\label{sec:related}
\paragraph{Parameter-space exploration.}
\Es{} and related methods are established alternatives to policy-gradient
RL~\citep{salimans2017es,sehnke2010pgpe,wierstra2014nes,maheswaranathan2019guided}.
\citet{salimans2017es} discussed the advantages of \Es{} on long horizons, and \citet{vemula2019contrasting} analyzed parameter
dimension, action dimension, and horizon. At LLM scale, \citet{esatscale2025} reported competitive \Es{}
performance, including gains over RL baselines on Countdown, and \citet{blessingdim} explained its scalability by showing that a small set of high-curvature directions dominates the improvement, which addresses the dimension cost but not when \Es{} should beat \Pg{}. An independently tuned \Pg{} run on the same Countdown setup reported rewards
that, averaged over the same mix of problems, exceeded those of both their methods~\citep{sana2026hpo}, which suggests that comparisons of the two methods
depend strongly on how the RL baseline is tuned.
Intrinsic-dimension and low-rank adaptation results provide further
context~\citep{li2018intrinsic,aghajanyan2021intrinsic,hu2022lora}. Our work adds a controlled
separation of weight sharing and credit assignment, trained comparisons with both methods tuned in the same
environment, and task horizons measured in the setup of each comparison.

\paragraph{Policy gradients and long-horizon agents.}
REINFORCE variants underpin current LLM RL~\citep{williams1992reinforce,shao2024grpo,deepseekr1,liu2025drgrpo}.
Finer credit assignment through Monte Carlo value estimates improved sample
efficiency~\citep{kazemnejad2024vineppo}. \citet{yue2025passk} found that several RL methods performed
similarly on reasoning and beat the base model only at small $k$ in pass@$k$.
Tmax and T1 addressed instability in terminal-agent training~\citep{ivison2026tmax,yang2026t1},
while TRACE estimated credit for individual tool actions~\citep{tao2026trace}.

\section{Discussion and future work}\label{sec:conc}
Terminal-Bench suggests that \Es{} can pull ahead of \Pg{} once tasks are long enough, and the natural next step is to
compare the two methods at equal cost. The analysis predicts that \Es{} needs fewer rollouts than \Pg{} beyond the crossover,
and the 7B benchmarks lie at or below its lower end. Measuring sample cost directly calls for a budget defined the same way for both methods, on a task whose effective horizon is in
the thousands of tokens, together with infrastructure that runs every perturbed model on full multi-turn
episodes. Such a comparison also needs care with the setup, since an apparent advantage of either method can easily come from
a mismatched environment or an untuned baseline. Whether such tasks involve few independent skills is a
further open question, because the reward-gradient count of skills fails on long tasks that the base model
rarely solves.

Multi-agent training is another promising direction to explore. When several LLM agents interact over many rounds,
the episode that produces a reward spans the turns of all agents, so its horizon can be far longer than that of
any single-agent task. Credit assignment also becomes much harder, because the reward depends on how the agents'
actions combine. Both factors raise the cost of \Pg{} but not that of \Es{}, which needs only the reward of the
whole episode. Whether \Es{} gains an advantage there depends on whether the joint behavior involves few
independent skills and whether the agents obtain the reward often enough to learn from it.

Beyond its variance, \Es{} also differs from \Pg{} in its exploration mechanism, which allows it to discover fundamentally different solutions. Because it explores in
parameter space, it follows the gradient of a smoothed reward, which repairs decisions on which the policy has
become confident in a wrong action, whereas the gradient that \Pg{} follows vanishes on them.
Training runs on the synthetic task in Appendix~\ref{app:dynamics} show this repair directly. \Es{} also needs no credit assignment and only forward passes, so it stores no
activations for backpropagation and uses less memory. With suitable infrastructure, every perturbed model can
run in parallel on inference engines, which may make \Es{} less demanding on hardware.

\section{Conclusion}\label{sec:final}
We asked when \Es{}, which explores in parameter space, can rival \Pg{}, which explores in action
space, for fine-tuning LLMs. Comparing the variance of the two estimators, we identified the regime in which \Es{}
should have the advantage. The task should involve few independent skills, because \Es{} pays for each of
them much more than \Pg{} does. Its effective horizon should be long, because only \Pg{} pays for the tokens
over which it must assign credit. Finally, the reward should carry a learning signal, so that either method
can learn at all. On real tasks at 7B, effective horizons are short to moderate and \Pg{} is ahead or tied on nearly
all of them, which places these benchmarks at the onset of this regime. Terminal-Bench gives a first indication of
what lies beyond it. On its far longer episodes, vanilla \Es{} with a stronger base model reaches a higher success rate
than a released \Pg{} checkpoint in a preliminary comparison, in line with the direction of this prediction.
Together, these results suggest that \Es{} will become a more competitive alternative as agentic tasks grow longer.

%% file: appendix_v2.tex
\appendix

\section{Derivation of the effective horizon}\label{app:tax}
This appendix derives the effective horizon used in Eq.~\ref{eq:pgdual} and states the assumptions it
needs. Consider one skill. The policy emits $T=k+F$ tokens. The first $k$ decision tokens determine the
reward, and the $F$ filler tokens do not. The reward $R$ is the fraction of decision tokens that match their
targets, so it lies in $[0,1]$. The \Pg{} estimate is $\hat g_{\Pg}=\sum_t\hat A_t\psi_t$, where
$\psi_t=\nabla_\theta\log\pi(a_t\mid s_t)$ is the score of the token at position $t$ and $\hat A_t$ is the
advantage assigned to it. The reward-relevant subspace $S$ is the row space of the Jacobian of the decision
logits with respect to $\theta$, over the decision positions of all skills, and $P_S$ projects onto it.

\paragraph{Assumptions.}
We make three assumptions, and all three hold by construction in the testbed. Tokens at different positions
are sampled independently given the parameters, the reward does not depend on the filler tokens, and the
advantage given to a filler token does not depend on that token.

\paragraph{Derivation.}
Every score has zero mean, because $\mathbb E[\psi_t]=\sum_a\nabla_\theta\pi(a)=\nabla_\theta 1=0$. Under the three assumptions, the token at a filler position $t$ is independent of everything else in any
covariance term that involves it. Every such term therefore contains the factor $\mathbb E[\psi_t]=0$. Filler therefore contributes no signal and no cross-covariance, only its own variance
$\mathbb E[\hat A_t^2]\,\mathbb E\lVert P_S\psi_t\rVert^2$. The decision terms do not depend on $F$, and we
normalize them so that each decision position counts as one. Dividing each filler term by the contribution of
one decision position gives
\begin{equation}\label{eq:taxfull}
\operatorname{tr}\mathrm{Cov}(P_S\hat g_{\Pg})\ \propto\ k+\sum_{\text{filler}}\beta_t\gamma_t\kappa_t,
\qquad
\beta_t=\frac{\mathbb E[\hat A_t^2]}{\mathbb E[\hat A_{\text{dec}}^2]},\quad
\gamma_t=\frac{\mathbb E\lVert P_S\psi_t\rVert^2}{\mathbb E\lVert\psi_t\rVert^2},\quad
\kappa_t=\frac{\mathbb E\lVert\psi_t\rVert^2}{\mathbb E\lVert\psi_{\text{dec}}\rVert^2}.
\end{equation}
In the testbed every position reads the parameters through a readout of the same scale, so $\kappa_t$ is the
ratio of the categorical score energies $1-\sum_v p_{t,v}^2$ of Eq.~\ref{eq:kappa}. When all positions
share the same factors, the sum reduces to $\Teff{}=k+\beta\gamma F$.

\paragraph{The credit and sharing factors.}
The credit factor $\beta_t$ measures how much advantage a filler token receives relative to a decision token.
Terminal credit gives every token the same advantage, so $\beta_t=1$. Per-token credit that gives filler no
advantage sets $\beta_t=0$. The sharing factor $\gamma_t$ is the fraction of the filler token's score
energy that falls inside $S$. The factor is zero when filler and decisions use disjoint parameters and equals $\dim S/D$ when the filler score
points in a random direction, as in the linear lift. In the shared MLP it depends on how similar the filler
inputs are to the decision inputs. With four skills we measure $0.57$ for local filler and $0.06$ for global filler (Table~\ref{tab:gates}). Filler therefore lengthens the effective horizon only when it receives credit, shares
directions with the decisions, and is uncertain.

\paragraph{Measured against predicted variance.}
Table~\ref{tab:gates} compares the reward-relevant variance measured at initialization on the shared MLP
with $(k+\gamma\beta F)/k$, with $\gamma$ measured on the same network. The prediction contains no fitted
parameter and agrees with the measurement to within $6\%$.

\begin{table}[htbp]
\centering\small
\caption{Reward-relevant \Pg{} variance relative to its value without filler, measured / predicted by
$(k+\gamma\beta F)/k$ ($r=4$ skills, $k=4$, $300$ gradient estimates of $64$ rollouts per condition, $3$
seeds). Here every estimate uses one fixed skill and is projected onto that skill's reward-relevant subspace,
so the no-filler variance contains no term from sampling the skill. Fig.~\ref{fig:blessing}b samples the skill
and projects onto the subspace of all four skills, so its growth with filler is smaller.}\label{tab:gates}
\begin{tabular}{lrrrr}
\toprule
condition & $\gamma$ & $F=64$ & $F=1024$ & $F=4096$ \\
\midrule
terminal credit, local filler & $0.57$ & $10.6$ / $10.1$ & $154.5$ / $145.8$ & $612.9$ / $580.2$ \\
terminal credit, global filler & $0.06$ & $2.0$ / $2.0$ & $17.0$ / $16.8$ & $64.8$ / $64.2$ \\
per-token credit, local filler & $0.00$ & $1.0$ / $1.0$ & $1.0$ / $1.0$ & $1.0$ / $1.0$ \\
\bottomrule
\end{tabular}
\end{table}

\paragraph{Scope.}
The derivation concerns the variance of one on-policy gradient estimate. The derivation does not cover bias from off-policy updates or trust regions, and it assumes that filler is truly
irrelevant to the reward, whereas in a language model early tokens can change later rewarded ones. Table~\ref{tab:duality} places the two
estimators side by side.

\begin{table}[htbp]
\centering\small
\caption{\textbf{The \Es{}--\Pg{} correspondence.} Both reward-relevant variances are a product of reward
noise, the energy per unit, and an effective count. In the testbed, $\Deff{}=r\cdot k$ does not change with
$D$, while $\Teff{}=k+\gamma\beta F$ grows with $F$ unless $\gamma$ or $\beta$ is zero.}
\label{tab:duality}
\begin{tabular}{lll}
\toprule
 & \Es{} \emph{(parameter space)} & \Pg{} \emph{(action space)}\\
\midrule
explores by & perturbing parameters $\theta\pm\sigma\varepsilon$ & sampling actions $a\sim\pi_\theta$\\
observes & scalar return $R$ & per-token scores $\psi_t$\\
samples per update & $P$ perturbations & $G$ rollouts\\
energy per unit & $1/\sigma^2$ (tunable) & $\kappa=1-\sum_v p_v^2\le1$\\
reward noise & $\mathrm{Var}(R)\le\tfrac14$ & $\mathrm{Var}(\hat A)\le\tfrac14$\\
nominal count & $D$ parameters & $T$ positions\\
effective count & $\Deff{}$ & $\Teff{}$\\
growth with nominal count & none & $\gamma\beta$ per added position\\
\bottomrule
\end{tabular}
\end{table}

\section{Testbed details and controls}\label{app:robust}
This appendix gives the details of the testbed and its training protocol, and the controls behind
Sections~\ref{sec:blessing} and~\ref{sec:phase}.

\paragraph{Task and policy.}
Every decision position of every skill has an input code drawn uniformly from the sphere of radius $4$ in
$512$ dimensions, and a target drawn uniformly from the $V=5$ tokens. The policy is a \texttt{tanh} network
with one hidden layer of width $512$ that maps a code to $V$ logits. A local filler code of skill $i$ is
$x_{\mathrm{fill}}=4\,\mathrm{normalize}(\hat c_i+\hat\xi)$, where $\hat c_i$ is the unit mean of the skill's
decision codes and $\hat\xi$ a fresh random unit vector. Its cosine is about $1/\sqrt2$ with $\hat c_i$ and
$1/\sqrt{2k}$ with each decision code. A global filler code is a random direction shared by all skills, with
cosine near zero to every decision code. Unless noted, $k=4$.

\paragraph{Tuning and evaluation.}
\Es{} replaces the returns by centered ranks and applies Adam to the resulting estimate, and \Pg{} applies
Adam to the REINFORCE gradient with a mean baseline. For every cell we run a grid on $4$ tuning seeds and select the configuration with the lowest mean cost,
where a failed run counts at its full budget. We evaluate the selected configuration on $8$ fresh seeds. The grid for \Es{} is a step size in $\{0.003,0.01,0.03,0.1,0.3\}$ and a perturbation scale
in $\{0.25,0.5,1,2\}$. The grid for \Pg{} is a step size in $\{0.0001,0.0003,0.001,0.003,0.01,0.03\}$ and an
entropy coefficient in $\{0,0.001,0.003,0.01,0.03,0.1\}$, with the entropy bonus normalized per position. A
tuning run stops once it has used four times the best mean cost found so far in its cell. A selected \Pg{}
configuration occasionally failed on one or two evaluation seeds, when filler noise locked a decision onto a
wrong token early, a failure that Appendix~\ref{app:dynamics} examines. For these cells, and for every cell of Fig.~\ref{fig:blessing}d,
we re-ran the six best configurations on $8$ (or $16$) tuning seeds, discarded any configuration with a
failed seed, and evaluated the selected one on $8$ (or $16$) fresh seeds. The parameter-count inset of
Fig.~\ref{fig:blessing}c uses four skills and scales the step sizes and the perturbation scale with one over
the square root of the width. If the criterion is instead $90\%$ of the decisions correct under the argmax,
the tie horizons are two to ten times longer, because \Pg{} reaches a correct argmax while still placing
much less probability on the targets than \Es{} does.

\begin{figure}[htbp]\centering
\includegraphics{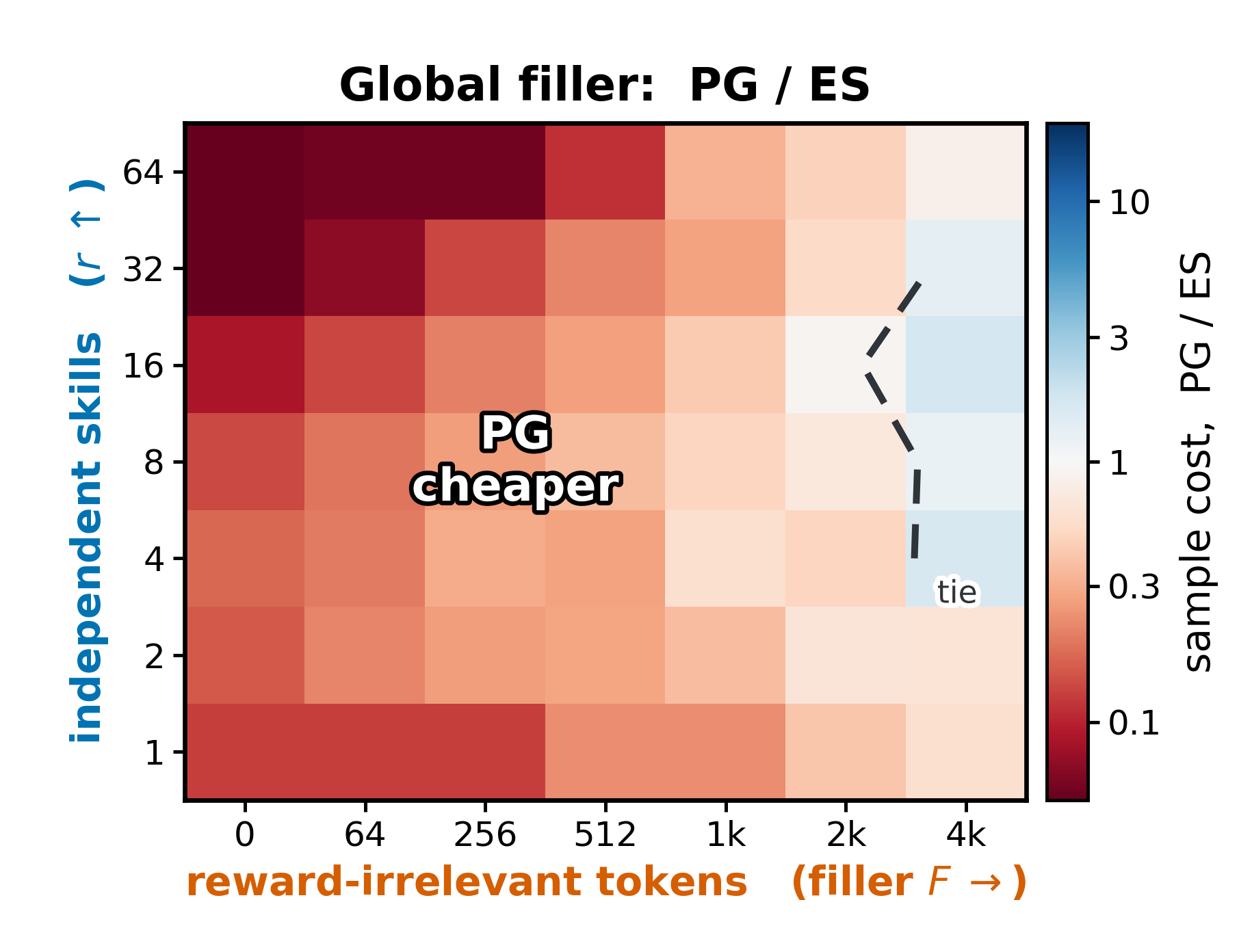}
\caption{\textbf{Global filler.} With filler that overlaps every skill, the cost of \Pg{} per skill grows with the
number of skills about as fast as that of \Es{}, so the two methods stay close throughout. Same grid and color scale as Fig.~\ref{fig:plane}a, with one set of filler
codes shared by all skills.}
\label{fig:plane-global}\end{figure}

\paragraph{The full grid.}
Table~\ref{tab:grid} lists the costs behind Figs.~\ref{fig:blessing}d and~\ref{fig:plane}.
Figure~\ref{fig:plane-global} shows the plane with global filler. A global filler score overlaps the decision directions of the visited skill by only $\gamma=0.06$ at every
number of skills. Its overlap with the directions of all skills together grows with $r$, reaching $0.61$ at
$64$ skills. Global filler therefore spreads noise over every skill, much as a parameter perturbation does,
and the two methods stay close across the plane.

\begin{table}[htbp]
\centering\small
\caption{Median rollouts per skill until the expected reward reaches $0.9$ ($k=4$). \Es{} does not depend on
the filler or on the credit scheme.}\label{tab:grid}
\begin{tabular}{rrrrrr}
\toprule
 & & \multicolumn{4}{c}{\Pg{}} \\
\cmidrule(lr){3-6}
skills $r$ & \Es{} & $F=0$ & $F=256$ & $F=1024$ & $F=4096$ \\
\midrule
\multicolumn{6}{l}{\emph{terminal credit, local filler}}\\
$1$ & $512$ & $64$ & $256$ & $1{,}408$ & $4{,}096$ \\
$2$ & $896$ & $144$ & $1{,}072$ & $1{,}536$ & $4{,}944$ \\
$4$ & $992$ & $176$ & $640$ & $2{,}712$ & $7{,}096$ \\
$8$ & $1{,}428$ & $200$ & $692$ & $2{,}756$ & $9{,}884$ \\
$16$ & $2{,}166$ & $186$ & $960$ & $2{,}468$ & $10{,}128$ \\
$32$ & $5{,}179$ & $253$ & $1{,}349$ & $3{,}599$ & $11{,}819$ \\
$64$ & $12{,}208$ & $359$ & $1{,}852$ & $5{,}742$ & $13{,}999$ \\
\addlinespace
\multicolumn{6}{l}{\emph{terminal credit, global filler}}\\
$1$ & $512$ & $64$ & $64$ & $128$ & $320$ \\
$2$ & $896$ & $144$ & $256$ & $352$ & $592$ \\
$4$ & $992$ & $176$ & $328$ & $608$ & $1{,}616$ \\
$8$ & $1{,}428$ & $200$ & $412$ & $752$ & $1{,}760$ \\
$16$ & $2{,}166$ & $186$ & $474$ & $982$ & $3{,}636$ \\
$32$ & $5{,}179$ & $253$ & $696$ & $1{,}550$ & $6{,}702$ \\
$64$ & $12{,}208$ & $359$ & $672$ & $4{,}331$ & $10{,}564$ \\
\addlinespace
\multicolumn{6}{l}{\emph{per-token credit, local filler}}\\
$1$ & $512$ & $64$ & $64$ & $64$ & $64$ \\
$4$ & $992$ & $152$ & $144$ & $144$ & $200$ \\
$16$ & $2{,}166$ & $182$ & $188$ & $186$ & $194$ \\
\bottomrule
\end{tabular}
\end{table}

\paragraph{Decisions per skill.}
We repeated part of the grid with $k=16$, for $1$ and $4$ skills, and compared it with the $k=4$ grid at the
same $r\cdot k$. The cost of \Es{} per skill agrees within $10\%$, as $\Deff{}=r\cdot k$ predicts. The tie horizon agrees within
a factor of about two in effective tokens, and less well in filler tokens.

\begin{figure}[htbp]\centering
\includegraphics{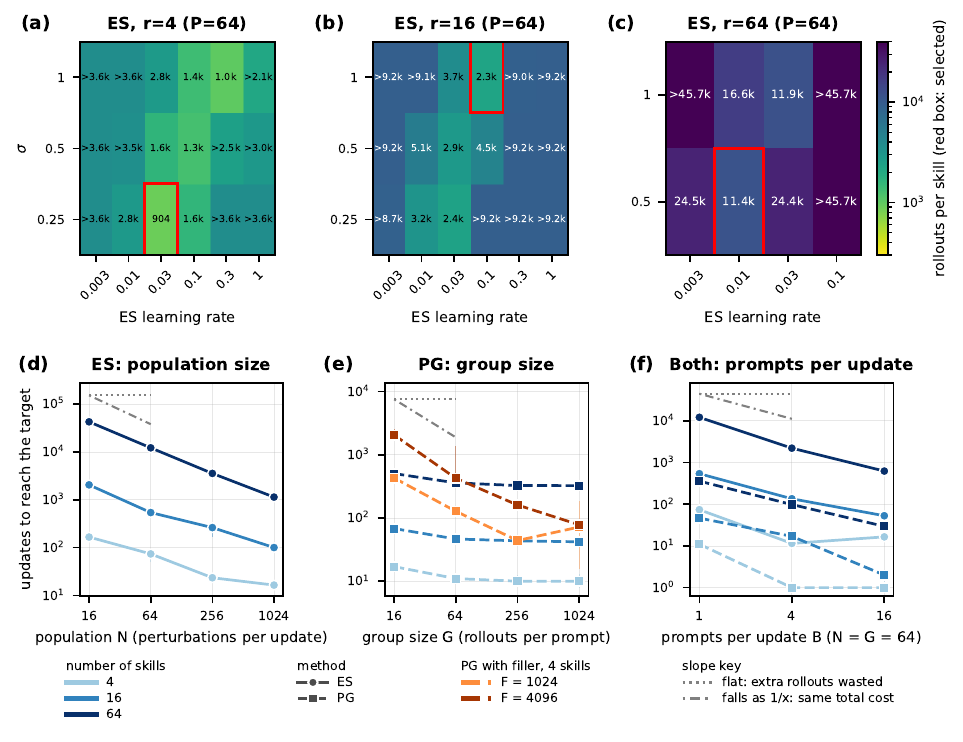}
\caption{\textbf{Hyperparameters and rollouts per update.} \Es{} must tune its step size and perturbation
scale together, and the number of rollouts per update barely moves where the two methods tie. All runs use $k=4$. \textbf{(a)--(c)} Cost of \Es{} in
rollouts per skill, the mean over $4$ tuning seeds, over its step size and perturbation scale, for $4$, $16$,
and $64$ skills. The red box marks the selected setting, and $>$ marks runs stopped at their tuning cap.
\textbf{(d)--(f)} Updates needed to reach the target, re-tuned at every value, against the population $N$ of
\Es{}, the group size $G$ of \Pg{} with and without filler, and the number of prompts per update with
$N=G=64$. A flat curve means that extra rollouts are wasted, and a curve falling as one over the rollouts per
update means that the total cost is unchanged.}
\label{fig:hparams}\end{figure}

\paragraph{Hyperparameters and rollouts per update.}
In Fig.~\ref{fig:hparams}a--c, the good settings of \Es{} lie along a band in which the step size grows with
the perturbation scale. The band moves with the number of skills, so \Es{} must be tuned in both. Smaller perturbations do not help. Figure~\ref{fig:hparams}d--f varies the rollouts per update,
re-tuning each method at every value. When the rollouts per update double, a method limited by its noise needs half as many updates. A method
limited by how far each update can move needs the same number, so the extra rollouts are wasted. Without
filler, \Pg{} is of the second kind because interference between skills limits its step, as Appendix~\ref{app:dynamics} shows, and smaller
groups are cheaper. With long filler its noise dominates, and
the largest group we tried is the cheapest. \Es{} is close to noise-limited over the populations we tried.
Spreading the rollouts of an update over several prompts changes the total cost of either method by less
than a factor of two. The fixed design of the main grid therefore understates the advantage of \Pg{} without
filler but barely moves the tie.

\paragraph{Other policy-gradient estimators.}\label{app:pgvariants}
We compare REINFORCE, RLOO \citep{ahmadian2024rloo}, GRPO \citep{shao2024grpo}, and
Dr.GRPO~\citep{liu2025drgrpo}, each with one advantage per trajectory. Their total variances differ by as much as
a factor of $100$, while the share of that variance in the reward-relevant subspace agrees within $5\%$. The effective horizon is
therefore the same for all four.

\section{Details on Terminal-Bench 2.1 results}\label{app:tbench}

The results in Section~\ref{sec:atlas} are those of \citet{schweighofer2026es}. They compare Qwen3.5-4B~\citep{qwen3.5} as the base model with the \Pg{} checkpoint trained on the same base model by \citet{ivison2026tmax} and with their \Es{} checkpoint. The \Pg{} checkpoint comes from the original version of the Tmax recipe.
Each checkpoint is evaluated five times on all Terminal-Bench 2.1 tasks using Amazon EKS as sandbox service, serving the LLMs through vLLM.
For all evaluations, the Vanillux harness from \citet{ivison2026tmax} was used, without the artificial cap on the number of interaction rounds per task.
The maximum sequence length was 64k with a maximum of 16k per turn.

For further details on the \Pg{} training setup, see \citet{ivison2026tmax}.
For \Es{}, the same training tasks (available as \texttt{Tmax-15K} through HuggingFace and Harbor-Hub) were used.
The reward is the binary correctness signal of the verifier scripts provided for each task.
For \Es{} training, the algorithm was used as implemented by \citet{esatscale2025} with population size = 32, $\alpha$ = 0.0005, $\sigma$ = 0.001, batch size = 256 as the main hyperparameters. 
Training was performed on an H200 cluster for a total of 200 iterations, selecting the checkpoint based on performance on a held-out set of 512 tasks of the Tmax dataset.

The two budgets are not directly comparable. Each \Es{} iteration evaluates 32 perturbations on 256 tasks, or 8,192 rollouts that need only forward passes. The released \Pg{} checkpoint comes from step 200 of a run with 8 prompts per rollout batch and 32 samples per prompt, so each step uses at least 256 rollouts, and active sampling and filtering add to that number by an amount that varies during training. A \Pg{} step also needs backward passes. An \Es{} iteration therefore uses more rollouts than a \Pg{} step, but each of those rollouts costs less to use, and we do not report a sample-cost comparison for this task.

\paragraph{Effective horizon of the Tmax setup.}
We measure the horizon of this setup with the probe of Appendix~\ref{app:probedef} on the base-model rollouts released with Tmax, which sample about $21{,}000$ tokens per episode. The references are the tool-call tokens of $74$ solved episodes on $24$ tasks. The horizon is about $7{,}300$, and about $19{,}000$ when the measured tasks are held out of the references. Resampling the reference tasks gives $2{,}500$ to $10{,}300$, so the horizon lies at the upper end of the crossover range or beyond under each variant. This is about sixty times the horizon of ALFWorld, the longest of the agentic tasks in Fig.~\ref{fig:measured}. For comparison, Qwen2.5-7B-Instruct in our own harness, where it solves almost none of the tasks, samples about $1{,}600$ tokens per episode and has a horizon of $318$, or $1{,}061$ with held-out references, on $62$ reference tasks.

\section{Learning dynamics, interference, and repair}\label{app:dynamics}
The costs of Sections~\ref{sec:blessing} and~\ref{sec:phase} summarize each training run by a single number. Here
we look inside the runs, recording after every update how much probability the policy puts on the target of
every decision. We call a decision \emph{locked} when a wrong token carries more than $95\%$ of its
probability.

\begin{figure}[htbp]\centering
\includegraphics{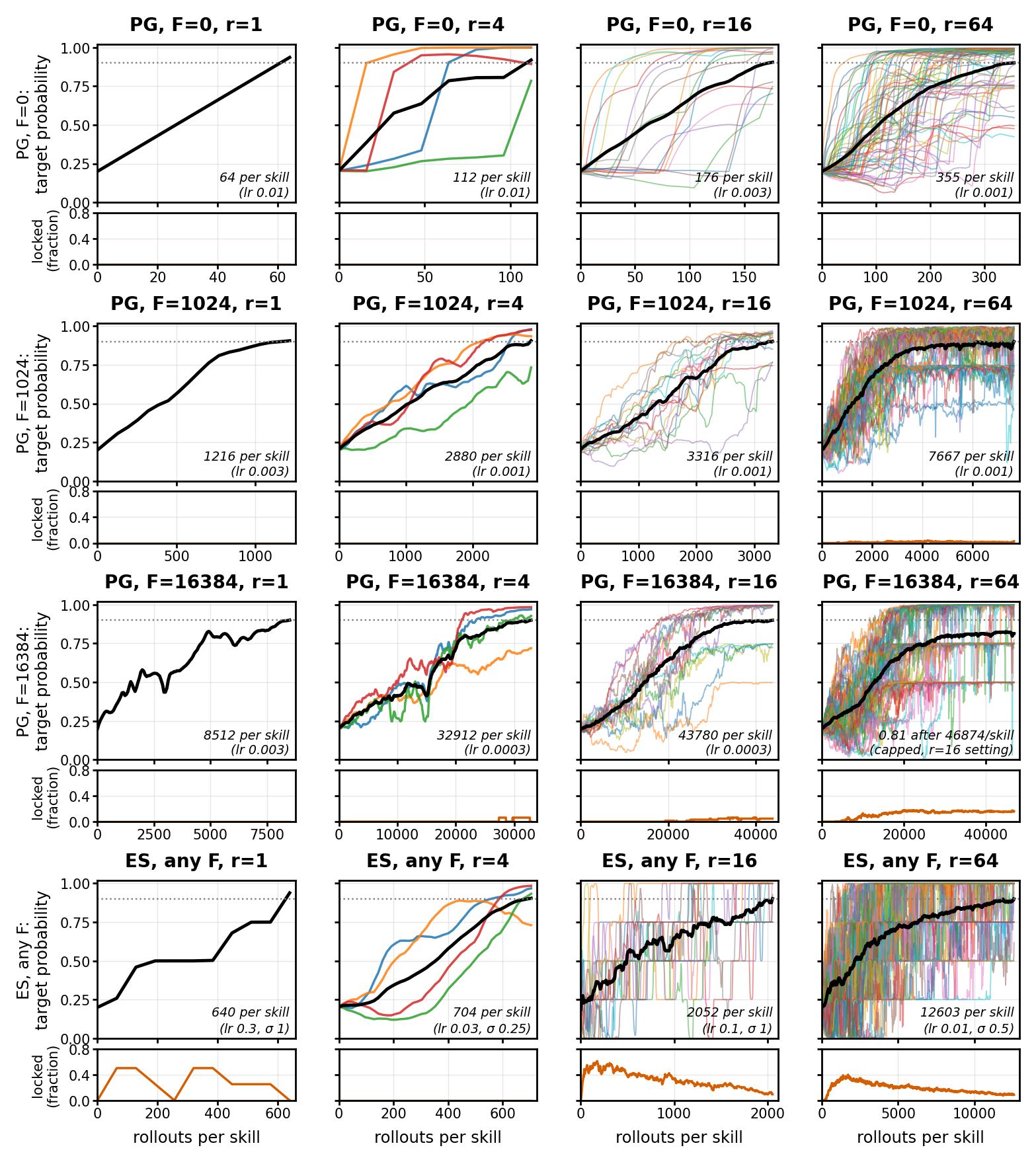}
\caption{\textbf{How each method learns the skills.} \Pg{} locks few decisions, mostly at long filler with many skills,
where its locks persist, while \Es{} commits decisions early and repairs most wrong ones later, and more skills slow
\Es{} down much more than \Pg{}. Each panel is one training run at the tuned setting of
the plane ($k=4$, local filler). Thin lines show, for each skill, the probability that the policy produces
its targets, and the black line is the average over skills. The strip below each panel shows the fraction of
decisions that are locked. Rows are \Pg{} with no filler, with $1{,}024$ filler tokens, and with $16{,}384$
filler tokens, and \Es{}, which filler does not affect. Columns are the number of skills. The number in each
panel is the cost in rollouts per skill. The \Pg{} run with $64$ skills and $16{,}384$ filler tokens was not
tuned separately and stops at its budget.}
\label{fig:dynamics-r}\end{figure}

\paragraph{Two ways of learning.}
\Pg{} learns gradually and locks few decisions (Fig.~\ref{fig:dynamics-r}). Its locks appear with long filler and
many skills, and they persist once formed. More skills
slow it down a little, and filler slows it down a lot. \Es{} learns in steps. It quickly commits many
decisions, some of them to wrong tokens, and corrects most of the wrong ones over the rest of the run. Filler
does not change this, but more skills slow \Es{} down much more than they slow \Pg{}. Committing early is not
what sets the cost of \Es{}, since settings that rarely lock cost about as much or more.

\paragraph{Why more skills slow \Pg{} down.}
All skills share one network, so an update for one skill also moves the decisions of the others. A large
step can push a decision of another skill onto a wrong token and lock it. With a single skill and no filler,
\Pg{} did not lock at any step size we tried, because its own updates move a decision toward its target. With
more skills, locks appear once the step is large enough, and almost all of them are created by the update of
another skill. The more skills share the network, the smaller the step at which locks appear, and the step
that tuning selects for \Pg{} shrinks accordingly. This smaller step is the main reason why the cost of \Pg{}
per skill rises with the number of skills. Filler can also lock a decision of a single skill, which causes the
occasional failed runs of Appendix~\ref{app:robust}.

\paragraph{How each method repairs a wrong decision.}
Let $m=\max_{v\ne\mathrm{tgt}}z_v-z_{\mathrm{tgt}}$ be the margin by which a wrong token leads in the logits
$z$ of a decision, so that the probability of the target is at most $e^{-m}$. A wrong decision is repaired if an update of its own skill pulls $m$ back toward zero. We write the average
pull as $\mu(m)$. For \Pg{}
with terminal credit, the expected gradient on the target logit is $p_{\mathrm{tgt}}\big(\mathbb E[R\mid
a=\mathrm{tgt}]-\mathbb E[R]\big)$, so $\mu_{\Pg}(m)\propto p_{\mathrm{tgt}}\le e^{-m}$. \Pg{} learns only from
tokens it samples, and once it is sure of a wrong token it no longer samples the target. For \Es{}, a
perturbation $\theta+\sigma\varepsilon$ changes the margin by about $\sigma\,\varepsilon^\top\nabla_\theta m$, a Gaussian with standard deviation $s=\sigma\lVert\nabla_\theta m\rVert$. The perturbed policy therefore
prefers the target with probability about $\Phi(-m/s)$. This probability changes appreciably with $m$ while
$m\lesssim s$, and rank shaping and Adam fix the size of each step. The pull $\mu_{\Es}(m)$ is therefore roughly
constant up to a margin that grows with $\sigma$. Beyond that margin \Es{} loses its signal as \Pg{} does.

\begin{figure}[htbp]\centering
\includegraphics{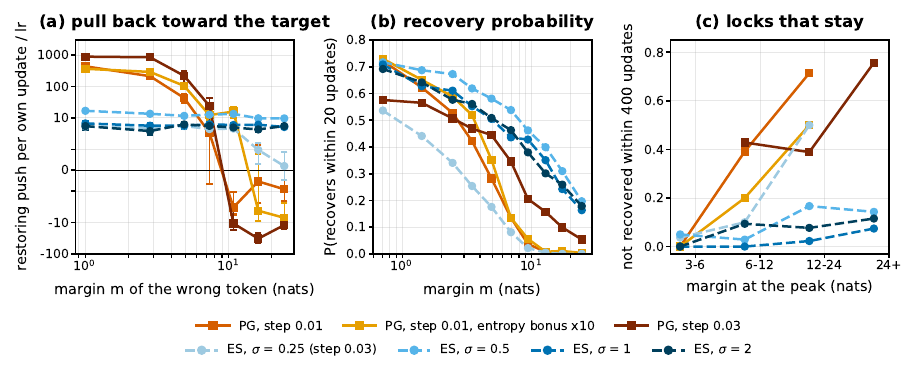}
\caption{\textbf{Repair of wrong decisions, measured in training runs.} \Es{} keeps pulling a wrong decision back toward its target up to a margin set by
$\sigma$, whereas the pull of \Pg{} vanishes as the wrong token gains a lead, so \Es{} repairs most locked
decisions and \Pg{} leaves many locked. All runs use $16$ skills, $k=4$, no filler, and four seeds per setting.
\Pg{} runs at step sizes above its tuned value of $0.003$, where it locks often enough to
measure. The horizontal axes give the margin $m$ by which a wrong token leads. \textbf{(a)} The
pull $\mu(m)$ toward the target in an update of the decision's own skill, divided by the step size. Values below zero mean that these updates on average push the wrong token further ahead,
and bars are standard errors. \textbf{(b)} Probability that the target is
back in the lead within $20$ updates. \textbf{(c)} Of the decisions locked when the number of locks is
largest, the fraction still not repaired $400$ updates later. Bins with fewer than five locked decisions are omitted.}
\label{fig:lock-theory}\end{figure}

\paragraph{What the measurements show.}
Figure~\ref{fig:lock-theory} confirms this picture at each method's own step sizes. The pull of \Pg{} is
strong when a wrong token leads by little and disappears within a few nats. The pull of \Es{} is weaker but stays roughly constant when $\sigma$ is large. With a small $\sigma$ it
disappears beyond a margin (panel a). Shallow locks are repaired by both methods, and the difference lies in deep ones (panel b).
Hundreds of updates later, roughly half of the deep locks of \Pg{} remain, whereas \Es{} with $\sigma\ge0.5$
has repaired most of them (panel c). Tuned \Pg{} therefore keeps its step small enough that deep locks do not
form, and tuned \Es{} can take steps that lock decisions because it repairs most of them.

\paragraph{Cost and final performance.}
These dynamics separate two questions that the variance analysis of the main text treats as one. Variance sets
how many rollouts a method needs to reach a solution, and therefore which method is cheaper when both reach the
same one. The exploration mechanism decides which solutions a method can reach at all. A decision that \Pg{}
has locked onto a wrong token rarely recovers, because \Pg{} no longer samples its target, whereas \Es{} perturbs
the parameters and still finds it. Where such locks form, the exploration mechanism rather than the variance
sets the final performance that a method attains.

\section{The benchmark probes and training comparisons}\label{app:deffcorr}
This appendix defines the probes behind Fig.~\ref{fig:measured}a, reports the checks we ran on them, and
describes the training comparisons of Fig.~\ref{fig:measured}b.

\subsection{Probe definitions}\label{app:probedef}
Both probes use the base model of the training run they are compared with, Qwen2.5-7B-Instruct for the
language, agentic, and Countdown tasks and Qwen2.5-Math-7B for the math tasks. They need a reference for every example, such as the answer, worked solution, or expert action sequence
provided with the benchmark. They also need rollouts sampled from the base model and backward passes, but no
reward model.

\paragraph{Reference gradients.}
All gradients are taken with respect to rank-$16$ LoRA adapters on the attention projections, the space in
which our \Pg{} runs train. For example $i$ with prompt $x_i$ and reference $y^\star_i$, the reference
gradient is
\begin{equation}\label{eq:probeloss}
g_i=\nabla_\theta\,\frac{1}{|y^\star_i|}\sum_{t}\log\pi_\theta\!\left(y^\star_{i,t}\mid x_i,y^\star_{i,<t}\right)\Big|_{\theta_0}.
\end{equation}
The span of the $g_i$ serves as the reward-relevant subspace $S$. Table~\ref{tab:atlasfinal} lists the number of directions they span, $\Deff{}=1/\bar\rho$, where $\bar\rho$ is
the mean of $\cos^2(g_i,g_j)$ over pairs of examples. On real tasks this is not a count of skills, for the
reason given under \emph{Reference style}.

\paragraph{Effective horizon.}
Let $\psi_t$ be the score vector of a sampled token and $P_S$ the projection onto $S$. The overlap of a set of
tokens with $S$ is $\sum_t\lVert P_S\psi_t\rVert^2/\sum_t\lVert\psi_t\rVert^2$, which we estimate without computing single score vectors through random-sign gradients $g_z=\nabla_\theta\sum_t z_t\log\pi_\theta(a_t\mid\cdot)$, which satisfy
$\mathbb{E}\lVert P_Sg_z\rVert^2=\sum_t\lVert P_S\psi_t\rVert^2$. We measure the overlap $\gamma_{\mathrm{roll}}$ of on-policy rollouts, four per example for $64$ examples, with
the prompt, generation limit, and sampling of the matching training run. We also measure the overlap
$\gamma_{\mathrm{unit}}$ of tokens sampled at the positions of the reference, with the reference as context. The estimate is
\begin{equation}\label{eq:teffprobe}
\Teff{}=\bar T\,\frac{\gamma_{\mathrm{roll}}}{\gamma_{\mathrm{unit}}},
\end{equation}
with $\bar T$ the mean number of generated tokens per rollout, and it assumes terminal credit. The generation
limit matters. Without the runs' limit and top-$k$ truncation the horizon of HellaSwag is $107$, against $35$
under the runs' own limit of $48$ tokens. The references of Countdown are solution expressions without the
reasoning that leads to them, so its horizon is probably underestimated relative to the math tasks.

\paragraph{Validation.}
On the testbed, where the true horizon is $k+\gamma F$, the estimator returns it with a median ratio of $0.97$ and a typical error of a factor of $1.4$ when the task has at least
four skills. With a single skill it is unreliable. All passes run in half precision. Repeating five tasks with
a more precise attention kernel changed $\Teff{}$ by at most $12\%$.

\paragraph{Held-out references.}
Rebuilding $S$ without the examples whose rollouts are measured changes the horizon by less than a factor of
$2.1$ on all but two tasks, SQuAD ($2.4$) and WebQuestions ($2.3$)
(Table~\ref{tab:atlasfinal}).

\paragraph{Noise of a \Pg{} update.}\label{app:pgnoise}
We also compare $\Teff{}$ with the noise of \Pg{} itself on the real model. For each task we form per-rollout gradients at the base policy, as the training runs compute them, on the fixed pool of training prompts. The noise-to-signal ratio of one update of six prompts is its variance divided by the squared norm of its expected value, with the signal estimated across prompts so that it carries no bias from a rollout's own noise. Across our $14$ tasks and the six tasks of prior work, in the setup of each training run, the rank correlation with $\Teff{}$ is $+0.83$ ($n=20$, $p<10^{-4}$), and within the $14$ tasks on which we also train both methods it is $+0.55$ ($p=0.04$). Much of the larger value comes from the separation between the math tasks and the rest. The main disagreements are classification tasks that the base model already answers well, whose small signal raises the ratio, and WebShop. The horizon estimate therefore orders tasks by \Pg{} difficulty, and it does not by itself predict which method wins a comparison.

\paragraph{Reference style.}
A human-written solution differs from the model's own writing in ways the reward does not care about. In a check with
Qwen2.5-7B-Instruct on math tasks with dataset solutions, one direction shared by all reference gradients
carries $20\%$ to $30\%$ of each of them and holds the count of a set of $256$ problems to a few dozen.
Replacing the dataset's solutions by correct solutions the model sampled itself removes the shared direction
and raises the count on MATH-500 from about a dozen to several hundred. The horizon of the same rollouts
changes less, rising by a factor of $2.8$ on MATH-500 and $2.6$ on GSM8K. We therefore read the horizon axis to
within about a factor of three.

\paragraph{Skill counts from reward gradients.}\label{app:skills}
The count that does return the number of skills uses the reward gradient of each example,
$\hat g_i=\frac{1}{G}\sum_g (R_g-b)\,\nabla_\theta\log\pi_\theta(\text{rollout}_g)$, with $b$ the mean reward of
the other rollouts. Sampling noise makes the gradients of different examples look orthogonal, so we form two
estimates $\hat g_i^A$ and $\hat g_i^B$ from disjoint halves of the rollouts and use
\begin{equation}\label{eq:skills}
\Deff{}=\frac{\big(\operatorname{mean}_i\,\hat g_i^A\!\cdot\hat g_i^B\big)^2}{\operatorname{mean}_{i\neq j}\,(\hat g_i^A\!\cdot\hat g_j^A)(\hat g_i^B\!\cdot\hat g_j^B)},
\end{equation}
in which every product pairs independent estimates. On the testbed with $256$ examples this returns the
number of skills from $1$ to $64$ for both filler types and every filler length, whereas a reference in a
fixed house style returns $1$ whatever the number of skills. Table~\ref{tab:skills} gives the count on eight
real tasks. The five short-answer tasks, sampled with a limit of $256$ new tokens, settle at $128$ rollouts
per problem, or $512$ on XSum, and give between $12$ and about $400$ skills. On MATH-500 the share of signal
in each gradient estimate is so small that settling would need about $1{,}000$ rollouts per problem. A problem
the model always or never solves has no reward gradient, so a task that the base model never solves cannot be measured
this way at any budget. On ALFWorld and WebShop, a rollout is one episode collected by the rollout loop of the training runs, with their step and token limits, sampling and reward, on games or goals from the training pool whose outcome varies in a screen of $16$ or $8$ rollouts. The gradient of an episode sums the log-probabilities of the agent's action tokens over all its steps. Leaving out the problems with the strongest signal changes every count by less than a factor of two, except on MATH-500 and on one of the $120$ WebShop goals, whose gradient is about $430$ times that of the median goal. With it, the WebShop estimate is $131$ ($19$--$299$), and the two halves give $315$ and $45$. Without it, the count stays between $18$ and $20$ whichever further goals are left out.

\begin{table}[htbp]
\centering\small
\caption{Number of independent skills from reward gradients, Eq.~\ref{eq:skills}, on Qwen2.5-7B-Instruct.
Problems $\times$ rollouts per problem; tokens per rollout; the estimate with its $68\%$ bootstrap interval over
problems; the estimates from the two halves of the rollout budget; the share of each gradient estimate that
is signal; and the reference-gradient dimension on the same problems. MATH-500 uses problems screened for a
mixed outcome in $8$ rollouts. $^{*}$Without the one goal of $120$ whose signal is about $430$ times the median.}\label{tab:skills}
\begin{tabular}{lrrlrrr}
\toprule
task & problems $\times$ rollouts & tokens & $\Deff{}$ (interval) & half budget & signal & reference \\
\midrule
\texttt{webq} & $64\times128$ & 117 & $12$ ($12$--$22$) & $15$, $11$ & $0.75$ & $60$ \\
\texttt{trivia} & $128\times128$ & 62 & $24$ ($20$--$36$) & $27$, $22$ & $0.64$ & $19$ \\
\texttt{squad} & $128\times128$ & 30 & $35$ ($27$--$53$) & $35$, $32$ & $0.88$ & $35$ \\
\texttt{translate} & $128\times128$ & 25 & $125$ ($103$--$247$) & $135$, $111$ & $0.88$ & $204$ \\
\texttt{xsum} & $64\times512$ & 43 & $385$ ($309$--$517$) & $453$, $325$ & $0.74$ & $14$ \\
\texttt{webshop}$^{*}$ & $119\times128$ & 85 & $19$ ($16$--$22$) & $22$, $16$ & $0.60$ & $30$ \\
\texttt{alfworld} & $64\times128$ & 141 & $128$ ($97$--$175$) & $95$, $151$ & $0.59$ & $41$ \\
\texttt{math-500} & $48\times192$ & 741 & $360$ ($63$--$1{,}011$) & -- & $0.15$ & $16$ \\
\bottomrule
\end{tabular}
\end{table}

\paragraph{Agentic episodes.}
For an agentic task, an example is one environment instance and its reference is the expert action sequence.
A rollout is one episode collected with the rollout loop of our training code, and the tokens of all its steps
are pooled. ALFWorld uses $256$ references from the simulator's planner. WebShop has no planner
and uses $44$ reward-reaching trajectories. The agents emit one short action per step.

\begin{table}[htbp]
\centering\small
\caption{Horizon and reference-gradient dimension of every task in Fig.~\ref{fig:measured}, each measured with
the model and sampling of its training run. Tokens are the mean generated tokens per rollout, the tick of
Fig.~\ref{fig:measured}a. Intervals are $68\%$ bootstrap intervals over examples. The held-out horizon leaves
the measured examples out of the reference subspace. The two $\Deff{}$ columns are the reference-gradient
dimension, unit-norm and weighted by gradient norm, and follow the style of the references rather than the
number of skills. Reward is the mean reward of the base model's rollouts. The three blocks are the agentic
tasks, the math tasks with Countdown, and the language tasks. $^{\dagger}$Fewer than $100$ references.
$^{\circ}$The reward varies across rollouts for fewer than $5\%$ of the examples.}\label{tab:atlasfinal}
\setlength{\tabcolsep}{4pt}
\begin{tabular}{lrrcrcrr}
\toprule
task & $N$ & tokens & $\Teff{}$ & held out & $\Deff{}$ & weighted & reward \\
\midrule
\texttt{alfworld} & 256 & 158 & $125$ ($114$--$140$) & $135$ & $41$ ($39$--$43$) & $49$ & $0.31$ \\
\texttt{webshop} & 44$^{\dagger}$ & 87 & $59$ ($50$--$66$) & $79$ & $30$ ($27$--$35$) & $28$ & $0.17$ \\
\midrule
\texttt{olympiad} & 256 & 1{,}202 & $522$ ($466$--$571$) & $560$ & $77$ ($72$--$83$) & $32$ & $0.10$ \\
\texttt{math-500} & 256 & 1{,}104 & $434$ ($341$--$527$) & $458$ & $39$ ($35$--$42$) & $6.9$ & $0.33$ \\
\texttt{aime} & 30$^{\dagger}$ & 1{,}037 & $380$ ($309$--$484$) & $558$ & $131$ ($102$--$178$) & $80$ & $0.07$ \\
\texttt{amc} & 28$^{\dagger}$ & 1{,}077 & $261$ ($191$--$362$) & $295$ & $101$ ($84$--$124$) & $64$ & $0.21$ \\
\texttt{minerva} & 69$^{\dagger}$ & 827 & $224$ ($176$--$299$) & $310$ & $47$ ($42$--$56$) & $45$ & $0.08$ \\
\texttt{countdown} & 256 & 577 & $143$ ($128$--$159$) & $142$ & $2.0$ ($2.0$--$2.0$) & $2.4$ & $0.29$ \\
\midrule
\texttt{hellaswag} & 256 & 48 & $35$ ($32$--$39$) & $40$ & $11.9$ ($11.5$--$12.6$) & $22$ & $0.20$ \\
\texttt{xsum} & 256 & 41 & $25$ ($22$--$27$) & $31$ & $15.4$ ($14.8$--$15.9$) & $16.0$ & $0.23$ \\
\texttt{translate} & 256 & 21 & $12.6$ ($10.7$--$14.4$) & $20.2$ & $213$ ($197$--$229$) & $314$ & $0.44$ \\
\texttt{webq} & 256 & 16 & $7.3$ ($5.5$--$9.2$) & $16.5$ & $59$ ($55$--$64$) & $54$ & $0.08$ \\
\texttt{squad} & 256 & 20 & $6.7$ ($4.9$--$8.3$) & $16.2$ & $35$ ($32$--$37$) & $20$ & $0.29$ \\
\texttt{trivia} & 256 & 16 & $5.5$ ($4.6$--$6.5$) & $6.2$ & $18.2$ ($17.2$--$19.4$) & $9.9$ & $0.11$ \\
\texttt{imdb} & 256 & 8 & $4.1$ ($3.8$--$4.5$) & $3.8$ & $1.6$ ($1.6$--$1.7$) & $1.6$ & $0.83$ \\
\texttt{winogrande} & 256 & 3 & $2.5$ ($2.4$--$2.6$) & $2.8$ & $225$ ($214$--$240$) & $183$ & $0.72$ \\
\texttt{agnews}$^{\circ}$ & 256 & 2 & $2.3$ ($2.2$--$2.4$) & $3.4$ & $13.8$ ($13.0$--$14.5$) & $29$ & $0.83$ \\
\texttt{boolq}$^{\circ}$ & 256 & 2 & $2.0$ ($1.8$--$2.3$) & $2.0$ & $2.7$ ($2.6$--$2.8$) & $4.0$ & $0.85$ \\
\texttt{arc}$^{\circ}$ & 256 & 2 & $1.8$ ($0.8$--$2.0$) & $1.0$ & $13.1$ ($12.3$--$13.9$) & $215$ & $0.88$ \\
\texttt{csqa} & 256 & 2 & $1.6$ ($1.2$--$1.8$) & $1.3$ & $27$ ($25$--$30$) & $1{,}086$ & $0.78$ \\
\bottomrule
\end{tabular}
\end{table}

\subsection{Training comparisons}\label{app:trained}
\paragraph{Setup.}
\Es{} perturbs all of the roughly $7.6$B weights of Qwen2.5-7B-Instruct with noise scale $\sigma=10^{-3}$. On
the language tasks, \Pg{} trains rank-$16$ LoRA adapters on the attention projections with Dr.GRPO. Both
methods train for $60$ updates and generate $72$ completions per update from a pool of $64$ training
examples. \Pg{} samples $12$ completions for each of six prompts. \Es{} scores the greedy completions of
each population member on the same prompts, either $12$ members on six prompts or $6$ members on $12$.

\paragraph{Training length.}
\Pg{} levels off within the $60$ updates: over the last $20$, its held-out score moves by at most $0.02$ on $10$ of the $12$ language tasks, and AG News and WinoGrande recover from an early dip. Training for $500$ updates on SQuAD, TriviaQA, WebQuestions, XSum, translation and HellaSwag ($2$ seeds for \Pg{}, $3$ for \Es{}, same settings) changes \Pg{}'s final score by at most $0.04$. At its population of $6$ to $12$, \Es{} does not gain from the longer runs: it ends below its base score or within $0.01$ of it on all six, and on SQuAD it peaks near $0.5$ before collapsing. A population of $30$ does help, reaching \Pg{}'s final score on SQuAD ($0.84$) after $150$ updates with about ten times \Pg{}'s samples (single seed, best of nine settings of step size and noise scale).

\paragraph{Tuning.}
We tune the learning rate of
\Pg{} and the step size and population split of \Es{}. On every task the same rule selects the configuration
with the best final score on the first seed. We report the mean and standard deviation over three seeds at that
configuration. Choosing the best checkpoint or the best single run instead leaves \Pg{} ahead or within
$0.03$ on every task. The selected setting often lies at an edge of the range searched, for \Pg{} on $11$ of the $14$ tasks and for \Es{} on $7$. The noise
scale of \Es{} and the group size and gradient clip of \Pg{} are not tuned.

\paragraph{Parameterization.}
To test whether low-rank adapters shape the comparison, we repeat all $12$ language tasks with \Pg{} training all weights. The data split, rewards, Dr.GRPO update, $60$ updates, and $72$ completions per update are unchanged. Master weights are stored in float32 under half-precision autocast, so that an update of the size of the learning rate is not rounded away, and we use 8-bit Adam because full-precision optimizer state for all weights does not fit in memory. We select the learning rate from $\{3\times10^{-7},10^{-6},3\times10^{-6},10^{-5},3\times10^{-5}\}$ by the same rule, and every selected rate lies inside this range. Table~\ref{tab:fullpg} gives the result. WinoGrande's full-weight runs differ widely between seeds.

\begin{table}[htbp]
\centering\small
\caption{\Pg{} on rank-$16$ LoRA adapters and on all weights, with \Es{} on all weights, on the language tasks at Qwen2.5-7B-Instruct. Final held-out scores, mean and standard deviation over three seeds at the selected configuration. The last column is the selected learning rate of full-weight \Pg{}.}
\label{tab:fullpg}
\begin{tabular}{lrrrrr}
\toprule
task & base & \Pg{} LoRA & \Pg{} all weights & \Es{} & rate \\
\midrule
SQuAD & $0.281$ & $0.820\pm0.022$ & $0.822\pm0.007$ & $0.378\pm0.097$ & $1\times10^{-6}$ \\
TriviaQA & $0.098$ & $0.566\pm0.008$ & $0.541\pm0.021$ & $0.099\pm0.020$ & $3\times10^{-6}$ \\
WebQuestions & $0.106$ & $0.300\pm0.016$ & $0.282\pm0.008$ & $0.107\pm0.004$ & $1\times10^{-6}$ \\
XSum & $0.260$ & $0.350\pm0.006$ & $0.341\pm0.004$ & $0.266\pm0.009$ & $3\times10^{-6}$ \\
translation & $0.521$ & $0.575\pm0.004$ & $0.581\pm0.009$ & $0.522\pm0.002$ & $3\times10^{-6}$ \\
IMDB & $0.880$ & $0.963\pm0.026$ & $0.978\pm0.031$ & $0.913\pm0.010$ & $3\times10^{-6}$ \\
AG News & $0.780$ & $0.823\pm0.070$ & $0.875\pm0.011$ & $0.803\pm0.013$ & $1\times10^{-6}$ \\
HellaSwag & $0.204$ & $0.222\pm0.005$ & $0.230\pm0.001$ & $0.207\pm0.001$ & $1\times10^{-6}$ \\
BoolQ & $0.865$ & $0.878\pm0.013$ & $0.882\pm0.010$ & $0.858\pm0.005$ & $1\times10^{-6}$ \\
CommonsenseQA & $0.840$ & $0.842\pm0.014$ & $0.840\pm0.004$ & $0.850\pm0.015$ & $1\times10^{-6}$ \\
ARC & $0.850$ & $0.853\pm0.005$ & $0.853\pm0.008$ & $0.852\pm0.012$ & $3\times10^{-6}$ \\
WinoGrande & $0.800$ & $0.802\pm0.033$ & $0.800\pm0.167$ & $0.783\pm0.005$ & $1\times10^{-5}$ \\
\bottomrule
\end{tabular}
\end{table}

\paragraph{Evaluation.}
Greedy completions are evaluated on $200$ held-out examples. Free-text answers are scored with normalized
token $F_1$, and letter and label answers must match in their first token. Table~\ref{tab:nlpfix} lists the
final scores. Across all $20$ comparisons of Fig.~\ref{fig:measured}b, the rank correlation between $\Teff{}$
and the difference in final score is $+0.20$. Without the six classification tasks it is $+0.74$ ($n=14$),
and $+0.52$ ($n=8$, $p=0.18$) on our own comparisons alone.

\begin{table}[htbp]
\centering\small
\caption{Head-to-head comparisons at Qwen2.5-7B-Instruct after tuning both methods, ordered by measured
horizon. Final held-out scores at each method's selected configuration, mean and standard deviation over three
seeds.}
\label{tab:nlpfix}
\begin{tabular}{lrrrrr}
\toprule
task & $\Teff{}$ & base & \Pg{} & \Es{} & \Es{} $-$ \Pg{} \\
\midrule
\texttt{csqa} & $1.6$ & $0.840$ & $0.842\pm0.014$ & $0.850\pm0.015$ & $+0.008$ \\
\texttt{arc} & $1.8$ & $0.850$ & $0.853\pm0.005$ & $0.852\pm0.012$ & $-0.002$ \\
\texttt{boolq} & $2.0$ & $0.865$ & $0.878\pm0.013$ & $0.858\pm0.005$ & $-0.020$ \\
\texttt{agnews} & $2.3$ & $0.780$ & $0.823\pm0.070$ & $0.803\pm0.013$ & $-0.020$ \\
\texttt{winogrande} & $2.5$ & $0.800$ & $0.802\pm0.033$ & $0.783\pm0.005$ & $-0.018$ \\
\texttt{imdb} & $4.1$ & $0.880$ & $0.963\pm0.026$ & $0.913\pm0.010$ & $-0.050$ \\
\texttt{trivia} & $5.5$ & $0.098$ & $0.566\pm0.008$ & $0.099\pm0.020$ & $-0.467$ \\
\texttt{squad} & $6.7$ & $0.281$ & $0.820\pm0.022$ & $0.378\pm0.097$ & $-0.443$ \\
\texttt{webq} & $7.3$ & $0.106$ & $0.300\pm0.016$ & $0.107\pm0.004$ & $-0.193$ \\
\texttt{translate} & $12.6$ & $0.521$ & $0.575\pm0.004$ & $0.522\pm0.002$ & $-0.052$ \\
\texttt{xsum} & $25$ & $0.260$ & $0.350\pm0.006$ & $0.266\pm0.009$ & $-0.085$ \\
\texttt{hellaswag} & $35$ & $0.204$ & $0.222\pm0.005$ & $0.207\pm0.001$ & $-0.015$ \\
\texttt{webshop} & $59$ & $0.404$ & $0.607\pm0.037$ & $0.525\pm0.007$ & $-0.082$ \\
\texttt{alfworld} & $125$ & $0.156$ & $0.486\pm0.079$ & $0.226\pm0.043$ & $-0.260$ \\
\bottomrule
\end{tabular}
\end{table}

\paragraph{Agentic tasks.}
Both methods train all weights and keep float32 master weights, \Pg{} under half-precision autocast. ALFWorld runs $30$ updates of $192$ episodes ($16$ games, each played by $12$ rollouts or population members) and WebShop $60$ updates of $128$ episodes ($8$ goals, $16$ each). Both methods train on the same games or goals in every update and are evaluated on fixed held-out games ($96$) or goals ($256$). The settings are selected by the same rule in runs one third as long: \Pg{}'s learning rate from $3\times10^{-7}$ to $1\times10^{-5}$ on ALFWorld and to $3\times10^{-5}$ on WebShop, and \Es{}'s step size from $1.25\times10^{-4}$ to $10^{-3}$ on ALFWorld and to $5\times10^{-4}$ on WebShop. Both tasks select $1\times10^{-6}$ for \Pg{} and $2.5\times10^{-4}$ for \Es{}. Rates of $1\times10^{-5}$ and above drive \Pg{}'s reward to zero, and on WebShop $3\times10^{-6}$ had a higher mean in the shorter runs ($0.58\pm0.04$) although seed $0$ selects $1\times10^{-6}$. Over the last third of training \Es{} gains $0.01$ on WebShop and $0.05$ on ALFWorld, and \Pg{} gains $0.04$ and $0.09$. \Pg{} is still rising on ALFWorld, so its score there is a lower bound. One \Es{} seed on ALFWorld is scored at its last evaluation (update $28$), because the environment crashed twice during its final update. On WebShop, a population of $30$ (\Es{} noise scale and step size reselected from $\sigma\in\{1,2,4\}\times10^{-3}$ and two step sizes, $8$ goals per update) scores $0.51\pm0.03$ after the same $7{,}680$ episodes and $0.51\pm0.04$ after $60$ updates ($14{,}400$ episodes), so a larger population does not move \Es{} on this task; \Pg{} reaches $0.57$ after $4{,}096$ episodes. In the shorter runs, \Pg{} reached $0.29\pm0.02$ on ALFWorld and $0.55\pm0.02$ on WebShop, or $0.22\pm0.03$ and $0.57\pm0.03$ with half-precision weights, and \Es{} $0.16\pm0.04$ and $0.50\pm0.01$. On WebShop the episode harness replaces a repeated action by a
step toward a purchase. The fitness of \Es{}, the training episodes of \Pg{}, and the evaluation all use it. The replacement matters, because
without it the base model's reward is several times lower.

\paragraph{What \Pg{} learns on ALFWorld.}
We repeated one \Pg{} run (seed $0$, final score $0.59$) and recorded every held-out episode. The base model mostly wanders: $83\%$ of its episodes run out of steps, and it picks up an object in $32\%$. After $30$ updates, $41\%$ run out of steps and $53\%$ pick up an object. The gain spreads over the task types at different times (Table~\ref{tab:alftypes}): placing an object improves first, cleaning from about update $14$, and placing two objects from about update $22$, while cooling and heating barely move. Each type needs its own behavior, so \Pg{} acquires several skills in turn, the regime in which \Es{} pays for each skill.

\begin{table}[htbp]
\centering\small
\caption{Held-out ALFWorld games won by \Pg{} (one seed) by task type, out of the number in brackets.}\label{tab:alftypes}
\begin{tabular}{lrrrr}
\toprule
task type & base & update $10$ & update $20$ & update $30$ \\
\midrule
pick and place ($18$) & $6$ & $10$ & $13$ & $13$ \\
pick two ($18$) & $4$ & $6$ & $8$ & $15$ \\
clean ($24$) & $2$ & $3$ & $10$ & $13$ \\
examine ($10$) & $2$ & $3$ & $7$ & $8$ \\
cool ($18$) & $1$ & $1$ & $0$ & $5$ \\
heat ($8$) & $1$ & $2$ & $3$ & $3$ \\
\midrule
all ($96$) & $16$ & $25$ & $41$ & $57$ \\
\bottomrule
\end{tabular}
\end{table}